\documentclass[english]{IEEEtran}
\usepackage[T1]{fontenc}
\usepackage[latin9]{inputenc}
\usepackage{verbatim}
\usepackage{float}
\usepackage{amsmath}
\usepackage{amssymb}
\usepackage{graphicx}

\makeatletter

\floatstyle{ruled}
\newfloat{algorithm}{tbp}{loa}
\providecommand{\algorithmname}{Algorithm}
\floatname{algorithm}{\protect\algorithmname}

\usepackage{bbm}
\usepackage{subfig}
\usepackage{graphicx}

\usepackage{hyperref}
\usepackage{adjustbox}
\graphicspath{{visualizations/}}

\usepackage{arydshln}

\usepackage{tikz}
\usetikzlibrary{shapes}
\pgfdeclarelayer{bg}    
\pgfsetlayers{bg,main}  
\usepackage{snapshot}

\usepackage{verbatim}

\@ifundefined{showcaptionsetup}{}{%
 \PassOptionsToPackage{caption=false}{subfig}}
\usepackage{subfig}
\makeatother

\usepackage{babel}
\begin{document}

\title{Modelling local-phase of images and textures with applications in
phase denoising and phase retrieval}

\author{Ido Zachevsky and Yehoshua Y. Zeevi}
\maketitle
\begin{abstract}
The Fourier magnitude has been studied extensively, but less effort has been
devoted to the Fourier phase, despite its well-established importance
in image representation. Global phase was shown to be more important
for image representation than the magnitude, whereas local phase,
exhibited in Gabor filters, has been used for analysis purposes in
detecting image contours and edges. Neither global nor local phase
has been modelled in closed form, suitable for Bayesian estimation.
In this work, we analyze the local phase of textured images and propose
a local (Markovian) model for local phase coefficients. This model
is Gaussian-mixture-based, learned from the graph representation of images, based on their complex wavelet decomposition. We demonstrate the applicability
of the model in restoration of images with noisy local phase and
in image retrieval, where we show superior performance to the well-known
hybrid input-output (HIO) method. We also provide a framework for
application of the model in a general setup of image processing.
\end{abstract}

\global\long\def\u#1{\underline{#1}}
\global\long\def\cov#1{\mbox{\ensuremath{\mathbb{C}}ov\ensuremath{\left\{  #1\right\}}}}
\global\long\def\E#1{\mbox{\ensuremath{\mathbb{E}\left\{  #1\right\}}}}
\global\long\def\P#1{\mbox{\ensuremath{\mathbb{P}\left\{  #1\right\}}}}
\global\long\def\var#1{\mbox{Var\ensuremath{\left[#1\right]}}}

\section{Introduction}

Inspecting images in the Fourier domain, one observes, as expected,
that the magnitude and phase ($A$ and $\phi$,
respectively) uniquely define an image. Inasmuch as there exists
a one-to-one correspondence from $A$ to the empirical autocorrelation, it follows that
higher moments, and specifically edge-type image skeleton structures,
are necessarily defined via the image phase, $\phi$. 

In natural stochastic textures (NST), an important facet of images that describes natural structure \cite{Zachevskya, Zachevsky2016}, we recognize a subset of cases
which is approximately Gaussian \cite{Zachevsky2016}, and is therefore
defined via its autocorrelation (second-order statistics). In this
subset, the phase is random. On the other hand, natural images contain
objects, separated from their background by edges and contours, that are dependent on their
phase coherence for edge representation. Some NST do share this property in
that their \emph{global} properties may be Gaussian, as expressed
via first and second order histograms, but they nevertheless have
local structures that are not suitable for a Gaussian or any second order model.

It is, therefore, desirable to find a model or description for the
\emph{local phase} of such natural ``structured stochastic'' textures. The
latter are defined as stochastic textures that incorporate phase structure. In such cases, applying noise or randomizing the phase will severely affect
the appearance of the image. For example, we observe (Fig. \ref{fig:phase_visu}) that the ``cameraman'' image (Fig. \ref{figw:pgt1}) is non-Gaussian (indicated by high kurtosis) and severely affected by phase distortions, whereas the stochastic image (Fig. \ref{figw:pgt26}) is closer to a Gaussian and is much less affected by the same phase distortion. The intermediate image (Fig. \ref{figw:pgt3}) also shows dependency on its phase, although it obeys Gaussianity, indicating that these type of images may benefit from a phase model.

The main contribution of this work is in providing a model for local phase coefficients; we review the importance of local phase, propose a model and method for estimating its parameters, and elucidate the usefulness of the proposed model by means of examples. An additional contribution is our novel phase retrieval algorithm, based upon our local phase model and the well-known HIO algorithm \cite{Fienup1982}. Since the proposed model can be used in Bayesian techniques, this is but one example of many other possible applications that can benefit from the proposed model.

\newcommand{\figim}[2]{\includegraphics[width=0.30\columnwidth,clip=false]{paper_visu/im#1_#2}}
\newcommand{\figimh}[2]{\includegraphics[width=0.34\columnwidth,clip=false]{paper_visu/im#1_#2}}

\newcommand{\kurta}[1]{\protect\input{paper_visu/im#1_K.txt}}
\newcommand{\subs}[1]{
\subfloat[Ground truth\label{figw:pgt#1}]{\figim{#1}{gt}} ~
\subfloat[Hist., $K=\protect\input{paper_visu/im#1_K.txt}$\label{figw:phist#1}]{\figimh{#1}{hist}}~
\subfloat[Phase distortion\label{figw:pdeg#1}]{\figim{#1}{deg}}
}
\begin{figure}
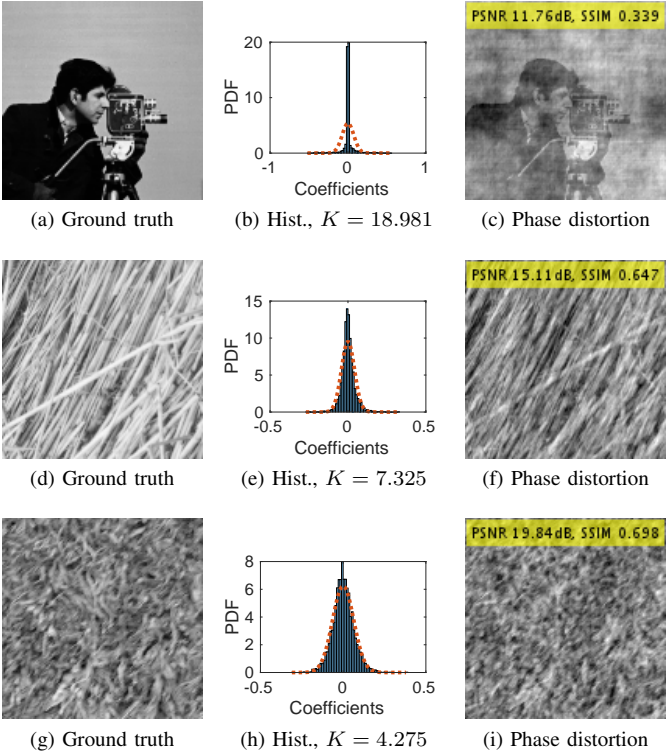
 \centering
\newcommand{\subreb}[2]{\protect\subref{figw:p#2#1}}

\subs{1} \\
\subs{3} \\
\subs{26}\\
\newcommand{\subreba}[1]{\subreb{1}{#1}, \subreb{3}{#1} and \subreb{26}{#1}}
\caption{Gaussianity and phase distortion: Images with Gaussian behavior are less affected by phase deviation, indicating that they are less dependent on their phase.  \subreba{gt}: Ground truth images. \subreba{hist}: Marginal histograms of wavelet coefficients with Gaussian fit (dashed red) and kurtosis value denoted by "K". \subreba{deg}: Result of phase distortion by AWGN with $\sigma=1$, for each of the ground truth images, respectively. PSNR and SSIM values indicated on the distorted image.
\label{fig:phase_visu}}
\end{figure}

\subsection{The local phase}

While a model for global phase is useful to some extent, in this work
we are concerned with the characteristics of local phase, defined
as the phase of a local structures. 
The local phase has several definitions, and it is either based on
a windowed Fourier transform or on the phase coefficients of complex
wavelet transforms or Gabor filters \cite{Porat1989,Behar1992}. There
are advantages in analyzing local phase, as we know that image reconstruction
via local phase yields better results compared with global phase-based
reconstruction \cite{Selesnick2005}, and, in fact, the local phase
portrays the basic image structure already in the first iterations
of the reconstruction \cite{Porat1989, Behar1992}. Further, local analysis is
computationally preferred in many cases, due to the simpler structures
exhibited locally, compared with the complexity of modelling entire images as a whole.

The local phase has been successfully applied in edge detection in
a method known as the phase congruency \cite{Kovesi1999}, which is
an edge detection method that incorporates the phase, normalized
by magnitude, thereby providing a detector which is invariant to various
degradations and illumination changes in the image. So have been the
zero crossings which are directly related to the local phase \cite{Rotem1986}.

The main property inherited by the phase is the local coherence of
different spatial frequencies, when they are in-phase (phase-locked).
This can be seen by the definition of the Fourier transform, as well
as by enforcing a constant magnitude to an image, observing that
edges are retained. This is due to the fact that image are known to
have a lowpass-type magnitude response, and enforcing a constant magnitude
serves as a highpass filter that emphasizes edges and contours. 

The coherence property is crucial in images since the magnitude energy
is not enough for image comprehension; edges and skeletons are of utmost importance
for image understanding and are defined by their coherence;
a known example is the simple step, which requires that all the frequencies will be phase locked, i.e. have the same
phase at the point of inflection and jump in intensity. 

\subsection{Modelling local phase}
In modelling phase we observe that unlike magnitudes (i.e. the spectrum)
that have a distinct exponential decay distribution, the phase distribution
appears at first to be inherently noisy or random, and is assumed in many
cases to be uniformly distributed. Consider a 2D magnitude
with several distinct high-energy coefficients that define the main
orientation of the patch. Other, less visible components, have lower
energy. 

The definition of the phase (as $\phi=\arctan2$$\left(\Im\hat{I},\Re\hat{I}\right)$
for $\hat I$, the Fourier transform of an image $I$) reveals that low energy
coefficients, that have low energy values, will vary significantly
in the range $\left[-\pi,\pi\right]$. Unlike the magnitude, in which
insignificant coefficients have values close to zero, their phases
will be approximately uniformly distributed. The incorporation of
these ill-defined coefficients will have counterproductive effect on
any model we seek. They therefore have to be scaled according to their
energy content.

Let us analyze an arbitrary patch, selecting only Fourier coefficients with high magnitudes.
If this patch depicts some coherent structure (e.g. an edge), its selected set of high-magnitude coefficients will exhibit structure in their phase. Locally, we assume that structures are simple enough and therefore can be described by steps, ramps or points. Consider
for example a patch that contains some oriented and uncentered edge. In the Fourier domain, this edge is described by some rotation and translation transformation of a centered and unoriented edge. 

These transform rules reveal the instability of the phase and consequent challenge in its modelling; while a
zero-mean, centered edge will be characterized by zero phase, a spatial
shift by one pixel will yield linear phase with slope corresponding
to the spatial translation distance.

In a densely scanned grid of patches, however, the same edge or otherwise
coherent structure will at one point appear at the center of a patch.
In the case of such a patch, the model is much simpler. Further, as
edges tend to exhibit locally only a single orientation, anisotropic
analysis is beneficial. Such an analysis is available in the form
of Gabor filters or complex wavelets, that generalize the Fourier
transform by incorporating scales and orientations in the image. 



One method that captures such coherence is the phase congruency \cite{Kovesi1999a}. The phase congruency is, however, method of analysis. It has been
used for edge detection \cite{Kovesi1999a}, segmentation, fusion
\cite{Singh2016} and other tasks. Further analysis methods have been
proposed for local phase, by proposing simple rules to link different
coefficients, or parametric distributions for marginal histograms
of phases \cite{Anderson2005,Anderson2005a,Miller2008,Miller2006,Wang2003}.
These methods were used as well as indicators for segment detection,
blur assessment and other tasks.

We, however, are interested in a complete model for the phase, that
can in turn be used in a Bayesian framework. In the sequel, we present
a graph- and wavelet-based image representation method and derive a
Markovian (based on a local neighborhood) phase model for textures
and images. This model is based on a mixture of Gaussians (GMM), which
describes the local phase coefficient, given its neighborhood in the
graph, by combining both spatial and scale relationships.

\subsection{Wavelets for phase processing}

Modelling both magnitude and phase can be accomplished by means of wavelet analysis. While
we are interested in phase models, we should use a wavelet family,
suitable for efficient processing in Bayesian frameworks with magnitude
models as well. Such a framework should possess several properties: efficient
inverse and forward transforms, orientational specificity and access
to phase of local coefficients. These requirements are translated
to using an orthogonal and discrete transform, an oversampled transform,
and a complex transform, respectively. 

We note that in the case of analysis (e.g. phase congruency),
the Gabor basis or the Morlet wavelets in their continuous versions
are used, as no inversion is required. A discrete, dyadic transform
emerges due to the requirement of a discrete and efficient transform.
Dyadic transforms are more limited in their ability to exhibit phase
responses, due to the sampling (decimation) of the coarser levels
in each iteration. Nevertheless, we show in the sequel how they can be used
for phase analysis.

A common property of discrete wavelet transforms is separability.
This property renders wavelets to be much easier to apply and invert,
but prevents orientational support. A wavelet family that satisfies
all aforementioned requirements is the dual-tree complex wavelet transform
(DTCWT) \cite{Selesnick2005}. It is a twice-oversampled, complex
coefficient, anisotropic wavelet transform. The DTCWT will be used in this work to model local phase.

\subsection{Related works}
Before presenting the graph-based model, we survey other phase-related studies. Local coherence has been investigated in \cite{Wang2003}, where
the authors analyze the phase of symmetric linear-phase complex wavelets
and show that given the self-similarity property in the Fourier domain:
\[
F\left(f/s\right)=K\left(s\right)F\left(f\right),
\]
where $s$ is the scale parameter and $F\left(t\right)$ is the coherent
feature in the transform domain, the phase of $F\left(t\right)$ in
high scales is equal to the phase of lower scales. The authors show that
this type of phase prediction and redundancy can be used for detecting
blur in images. Blur, while not disruptive to global phase (for a
large family of symmetric, zero-phase blur filters), does distort
the local phase correspondence, as expressed by analyzing the
local phase in different scales.

Direct application of the conclusions in \cite{Wang2003} requires
the application of a specific wavelet family; the authors used the complex
steerable wavelets with linear phase, i.e. the wavelet function is
a rotation of a prototype function, $f_{s}\left(t\right)=g\left(t\right)\cdot e^{-j\omega_{c}s}$,
where $g\left(t\right)$ is the prototype symmetric low-pass filter
with cut-off frequency $\omega_{s}$. The linear phase property is
required for this analysis. 

In \cite{Vo2010}, an analysis of the phase was performed for three
wavelet families, the DTCWT as well as the pyramidal dual-tree directional
filter bank (PDTDFB) and the uniform discrete curvelet transform (UDCT).
Here, the authors analyze the relative phase and propose statistical
models for it. The authors fit different distributions for the marginal
and joint histograms and use the distribution's parameters in characterising
different textures. 

The phase in the case of DTCWT, of interest in the context of our
work, was analyzed in \cite{Anderson2005a} by means of the so-called
inter-coefficient product (ICP), which analyzes phases of adjacent
scales. The ICP is a transform based on coefficient phase difference
that achieves translation invariance in detecting orientational features
in images. While it is an invertible transform, it is non-orthogonal.
Thus, small changes in coefficients may propagate in an undesirable manner
to the image space. 

In \cite{Miller2006,Miller2008}, the authors model phase relationship
of adjacent complex wavelet scales to perform denoising. They enforce
a certain property of the phase to yield more visually pleasing structures.
It is worth noting that many applications of the phase (e.g. \cite{Singh2015,Singh2016})
have to do with using the edge map from phase congruency for segmentation,
registration or other tasks that need the skeleton structure. 

We also note the use of quaternion-based methods, that extend beyond
complex numbers and provide further meanings to phases and magnitudes
\cite{Bayro-Corrochano2010}. These methods are, however, beyond the scope of
this paper.

\section{Wavelet decomposition as a graph } \label{sec:decomposition-graph}

\renewcommand{\figim}[2]{\includegraphics[width=0.23\columnwidth,clip=false]{coeff_visu/#1_#2}}

\renewcommand{\subs}[1]{
\subfloat[Ground truth\label{figw:gt#1}]{\figim{#1}{gt}} ~
\subfloat[$80\%$ rand. local phase\label{figw:rec#1}]{\figim{#1}{rec}}~
\subfloat[Rand. phase locations\label{figw:coeffs#1}]{\figim{#1}{coeffs_o2_i1}}~
\subfloat[$80\%$ rand. global phase\label{figw:recg#1}]{\figim{#1_global}{rec}}
}
\begin{figure}
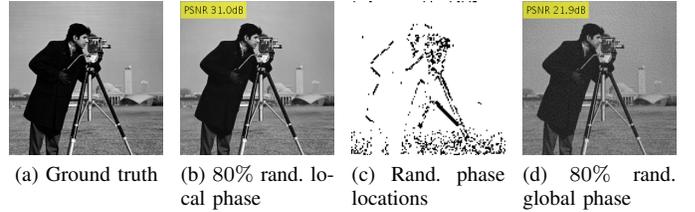
 \centering
\newcommand{\subreb}[2]{\protect\subref{figw:#1#2}}

\subs{cameraman}

\caption{Effect of phase randomization: Randomizing $80\%$ of DTCWT lowest-energy phase coefficients does not visually affect the image \subreb{rec}{cameraman}, unlike randomization of global phase \subreb{recg}{cameraman}. The randomized local coefficients belong to non-structured segments \subreb{coeffs}{cameraman}. \label{fig:neg-coeffs}}
\end{figure}

When modelling phase, it is important to separate the wheat from the
chaff, since low-energy coefficients that do not reflect structure
will affect estimated statistical models. We observe that keeping
only high-energy phase coefficients, while randomizing low-energy ones,
in the DTCWT domain, has negligible effect on the visual appearance
of images. We, therefore, assume that the low-energy phase is less important.

In Fig. \ref{fig:neg-coeffs}, for instance, we observe that randomizing $80\%$
of the low-energy local phase components (in the DTCWT sense) has
negligible effect on the image. Further, analyzing the randomized
local phase coefficients, we observe that, as expected, most of the
phase coefficients are discarded at non-edge points. On the other
hand, keeping the same percentage of coefficients ($20\%$) for the
global phase, results in much different behaviour with visible artefacts
(Fig. \ref{fig:neg-coeffs}). 

\begin{figure}
\centering\begin{adjustbox}
{max totalsize={1\columnwidth}{1\textheight},center}
\begin{tikzpicture}[scale=0.2]
\tikzstyle{every node}+=[inner sep=0pt]

\node (C3) [draw,ellipse,fill=white!90!blue] at (40.5,-30.5) {$c^{i}_{x-1,y}$};
\node (C4) [draw,ellipse,fill=white!90!blue] at (49.3,-21.3) {$c^{i}_{x+1,y}$};
\node (C0) [draw,circle,fill=white!90!blue] at (44.8,-25.9) {$c^{i}_{x,y}$};
\node (C1) [draw,ellipse,fill=white!90!blue] at (52.3,-25.9) {$c^{i}_{x,y+1}$};
\node (C2) [draw,ellipse,fill=white!90!blue] at (36.5,-25.9)  {$c^{i}_{x,y-1}$};

\node (D) [draw,diamond, inner sep=0pt,fill=white!90!blue] at (44.8,-12.4) {$c^{i-1}_{x,y}$};
\node (E1) [draw,rectangle,fill=white!90!blue] at (50.5,-40.5) {$c^{i+1}_{x+1,y}$};
\node (E2) [draw,rectangle,fill=white!90!blue] at (44.8,-45.5) {$c^{i+1}_{x,y+1}$};
\node (E3) [draw,rectangle,fill=white!90!blue] at (38.9,-40.5) {$c^{i+1}_{x,y-1}$};
\node (E4) [draw,rectangle,fill=white!90!blue] at (32.2,-45.5) {$c^{i+1}_{x-1,y}$};

\path [->] (C0) edge (C1);
\path [->] (C0) edge (C2);
\path [->] (C0) edge (C3);
\path [->] (C0) edge (C4);
\path [->] (C0) edge (D);
\begin{pgfonlayer}{bg}
\node[trapezium, draw, minimum width=3cm,
trapezium left angle=60, trapezium right angle=120, trapezium stretches=true,minimum height=1.5cm, minimum width=5cm, fill=white!90!green]
    at (45,-12.5) [inner ysep=1pt,inner xsep=1pt] {};
\node[trapezium, draw, minimum width=3cm,
trapezium left angle=60, trapezium right angle=120, trapezium stretches=true,minimum height=2.7cm, minimum width=8cm, fill=white!90!green]
    at (44,-26) [inner ysep=1pt,inner xsep=1pt] {};
\node[trapezium, draw, minimum width=3cm,
trapezium left angle=60, trapezium right angle=120, trapezium stretches=true,minimum height=3.3cm, minimum width=10cm, fill=white!90!green]
    at (43,-43) [inner ysep=1pt,inner xsep=1pt] {};

	\path [->] (C0) edge (E1);
	\path [->] (C0) edge (E2);
	\path [->] (C0) edge (E3);
	\path [->, bend left=0] (C0) edge (E4);

\end{pgfonlayer}
\node at (22,-50) {$l_{i+1}$};
\node at (27,-31.5) {$l_i$};
\node at (36,-15.2) {$l_{i-1}$};
\end{tikzpicture}\end{adjustbox}

\caption{Local graph representation of a wavelet coefficient $c_{x,y}^{i}$
on level $i$ and at spatial location $\left(x,y\right)$. Ellipses represent
spatial translations, rectangles represent finer levels and the diamond
represents the coarser level. $l_{i}$ denotes the $i$th decomposition
level. \label{fig:A-local-graph}}
\end{figure}
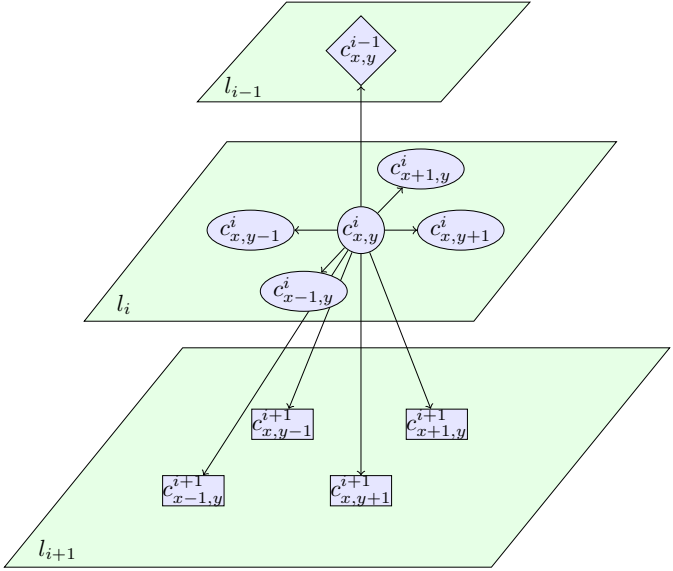

We represent an image by a graph as follows (Fig. \ref{fig:A-local-graph}, where the neighbors of a single coefficient, $c_{x,y}^{i}$, are depicted for a decomposition at level $i$ with spatial location $\left(x,y\right)$):
each wavelet coefficient entry has at most $9$ neighbors: $4$ spatial
neighbors (in a 4-neighborhood system), $4$ children that belong
to the finer decomposition level and $1$ parent that belongs to the
coarser level. 

In DTCWT, each decomposition level encodes information in $6$ spatial orientations that provide, in turn, more details and orientational information than a standard discrete wavelet transform. The choice of $6$ orientations rather than any other, arbitrary, number originated in vision research and was previously adopted in various schemes of image processing and computer vision \cite{Porat1988}. 

Each node in the graph contains the magnitude and phase
of all $6$ orientations. Nodes on the graph's boundary that contain
less than $9$ neighbors are discarded in our analysis. 

The exact neighborhood structure is derived from the wavelet transform used in the graph decomposition. The proposed graph structure depicted in Fig. \ref{fig:A-local-graph} has 4 children per node, that provide finer details in 4 projections of the same spatial location. Other wavelet transforms may yield different graph decomposition structures, but the same principal will be retained. We use the term ``neighbors'' as it is commonly used to denote neighbors in a graph, but we note that while neighbors at the same level (e.g. Fig. \ref{fig:A-local-graph}, level $l_i$) belong to adjacent spatial locations, neighbors in adjacent levels (e.g. $l_{i-1}$, $l_{i+1}$) have the same spatial location, but different wavelet frequency bands.

As an example of possible use of the graph, let us consider paths
of coarse-to-fine coefficients (i.e. a vector in the size of the transform
depth). Each such path starts from the finest coefficient and finishes
at the coarsest level, and for each finest coefficient we have 6 originating
paths, one for each orientation. In this case, we consider a \emph{low-energy
path}; i.e. a path with average coefficient magnitude lower than a certain
threshold.

Before resorting to complete modelling, we would like to gain first
some intuition on the graph structure. It is interesting to note that
randomization of phases in low-energy paths yields smoothing of fine
details, while retaining large-scale features (Fig. \ref{fig:rand-path}). 

\renewcommand{\figim}[1]{\includegraphics[width=0.235\columnwidth,clip=false]{coeff_visu/wood_paths_#1}}

\renewcommand{\subs}{
\subfloat[Ground truth\label{figw:gt}]{\figim{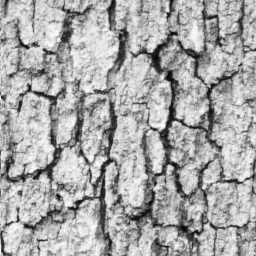}} ~
\subfloat[$29\%$ rnd.~pha. \label{figw:rec_1}]{\figim{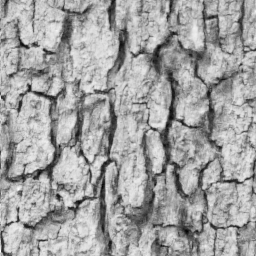}}~
\subfloat[$76\%$ rnd.~pha. \label{figw:rec_2}]{\figim{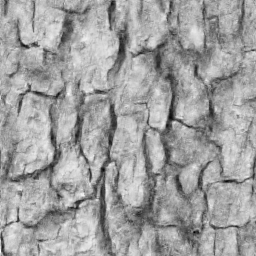}}~
\subfloat[$97\%$ rnd.~pha. \label{figw:rec_4}]{\figim{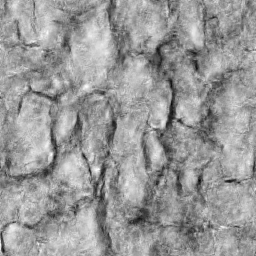}}
}
\begin{figure} \centering
\newcommand{\subreb}[1]{\protect\subref{figw:#1}}

\subs

\caption{Phase randomization of DTCWT paths: Images with randomized phase paths are shown with ascending extent of randomized content. Images in \subreb{rec_1}, \subreb{rec_2} and \subreb{rec_4} depict results of phase randomization of $29\%$, $76\%$ and $97\%$ of the coefficients, respectively. The images with  randomized phase appear natural due to randomness of complete paths. \label{fig:rand-path}}
\end{figure}

Analyzing the phase for high- and low-energy paths reveals a different
distribution for each case. We observe that in the high magnitude
paths, the marginal cdf of phase coefficients is not uniform, unlike
the low-magnitude paths (Fig. \ref{fig:Cdf-of-marginal}). This was
evaluated on the ``cameraman'' image. The same behavior was observed
in other images.

\newcommand{\curww}{0.32}
\newcommand{\curo}{1}
\newcommand{\addimga}{\includegraphics[width=\curww\columnwidth]{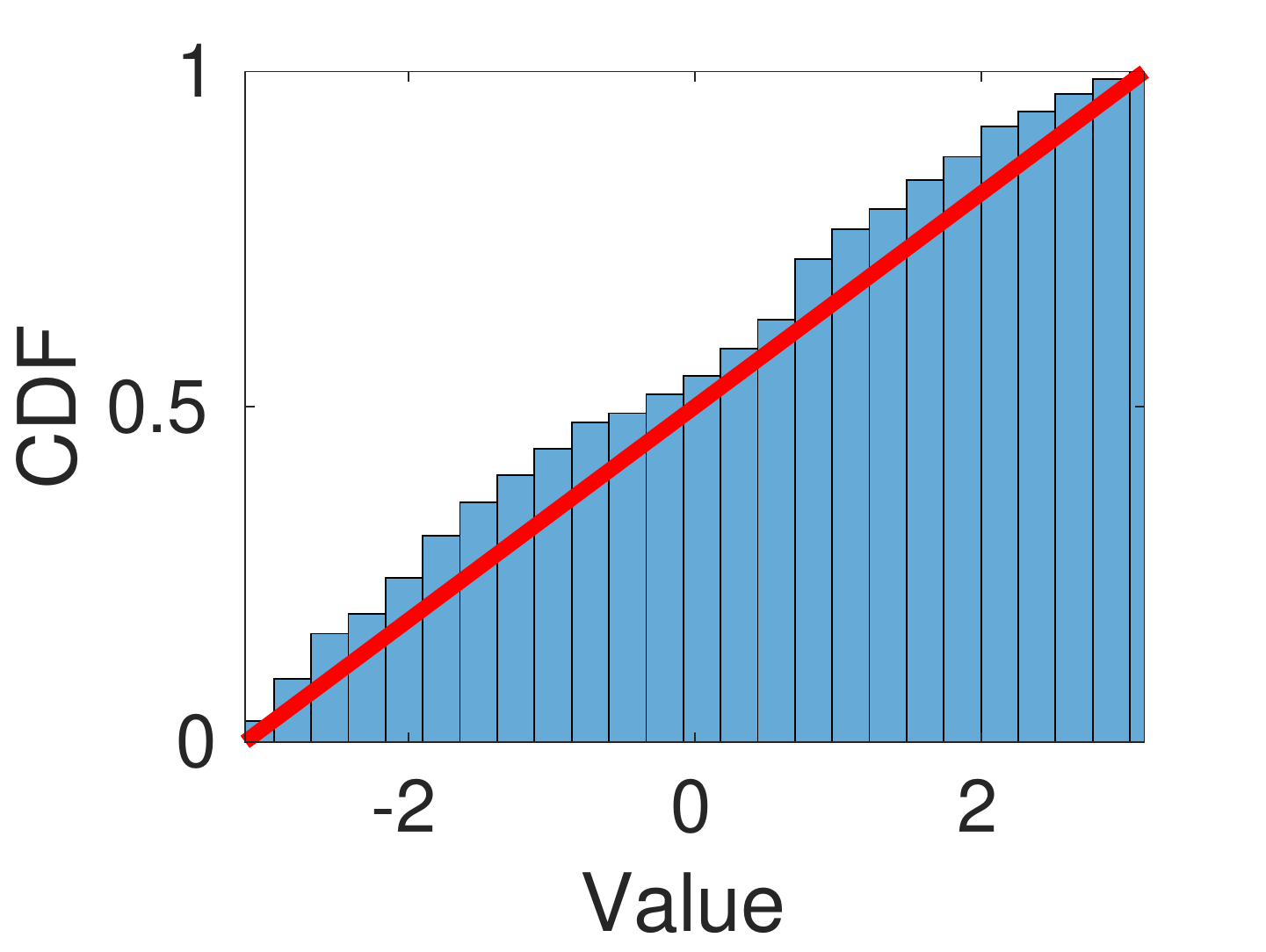}}
\newcommand{\addimgb}{\includegraphics[width=\curww\columnwidth]{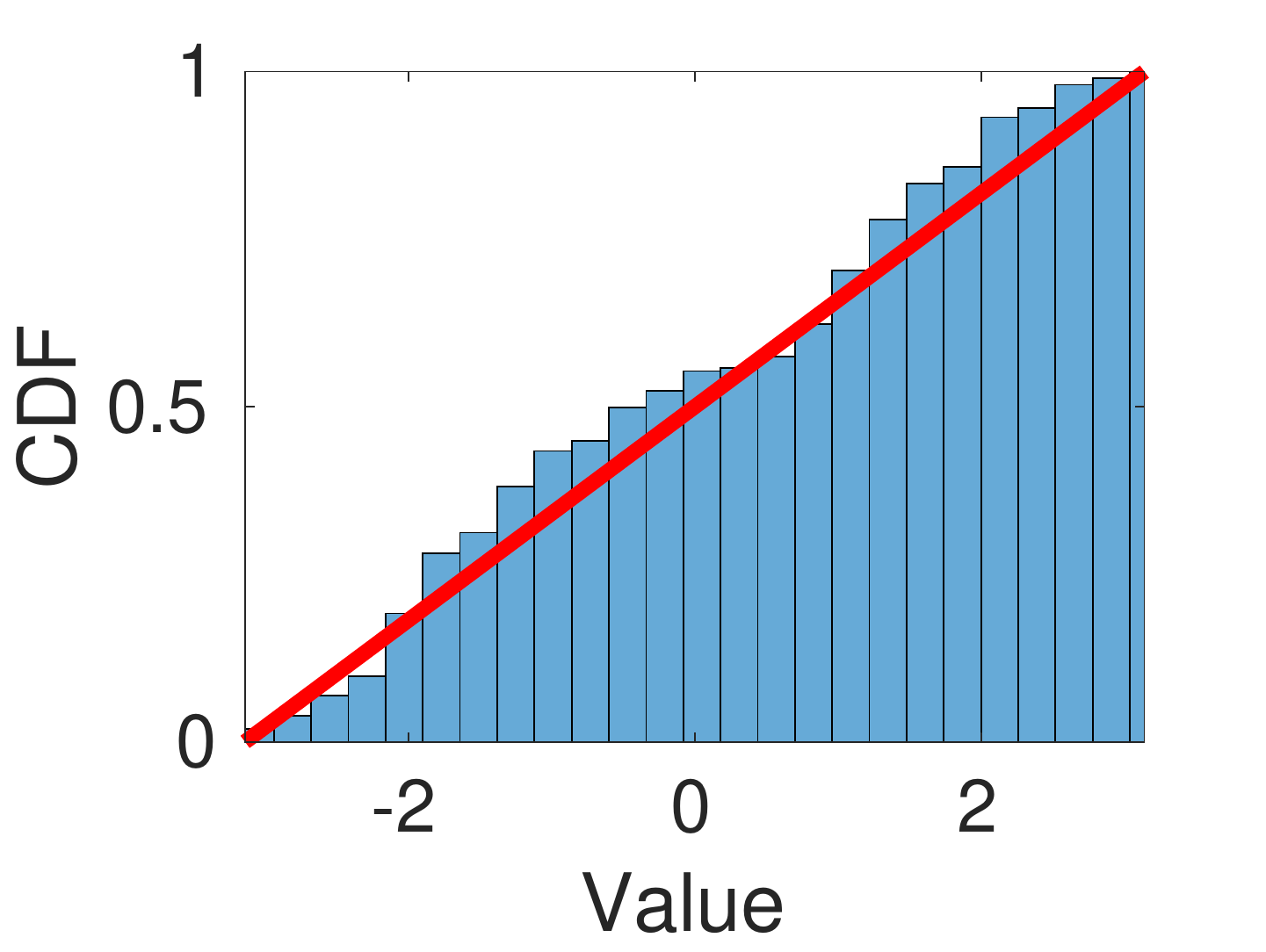}}
\newcommand{\addimgc}{\includegraphics[width=\curww\columnwidth]{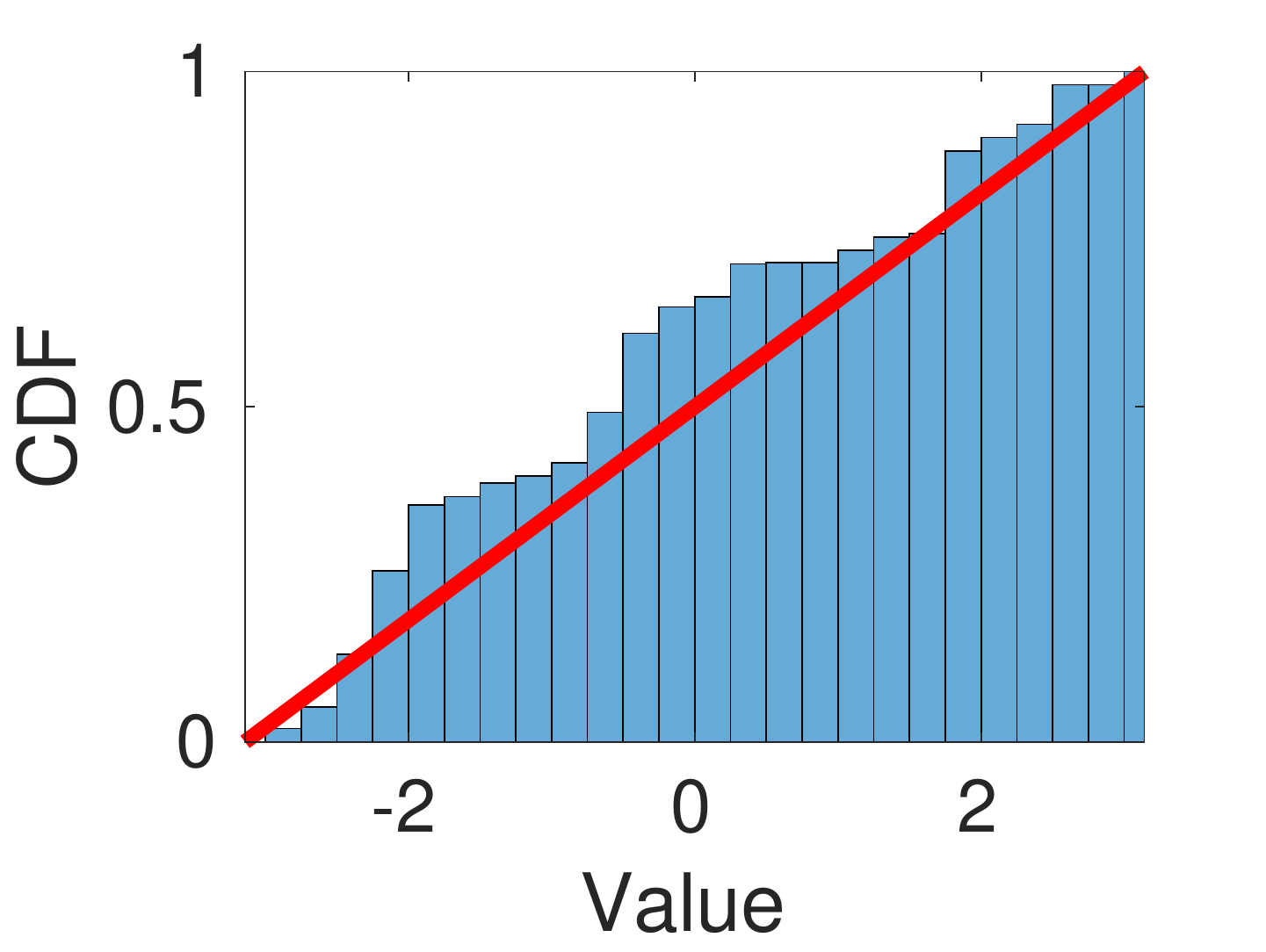}}
\newcommand{\addimba}{\includegraphics[width=\curww\columnwidth]{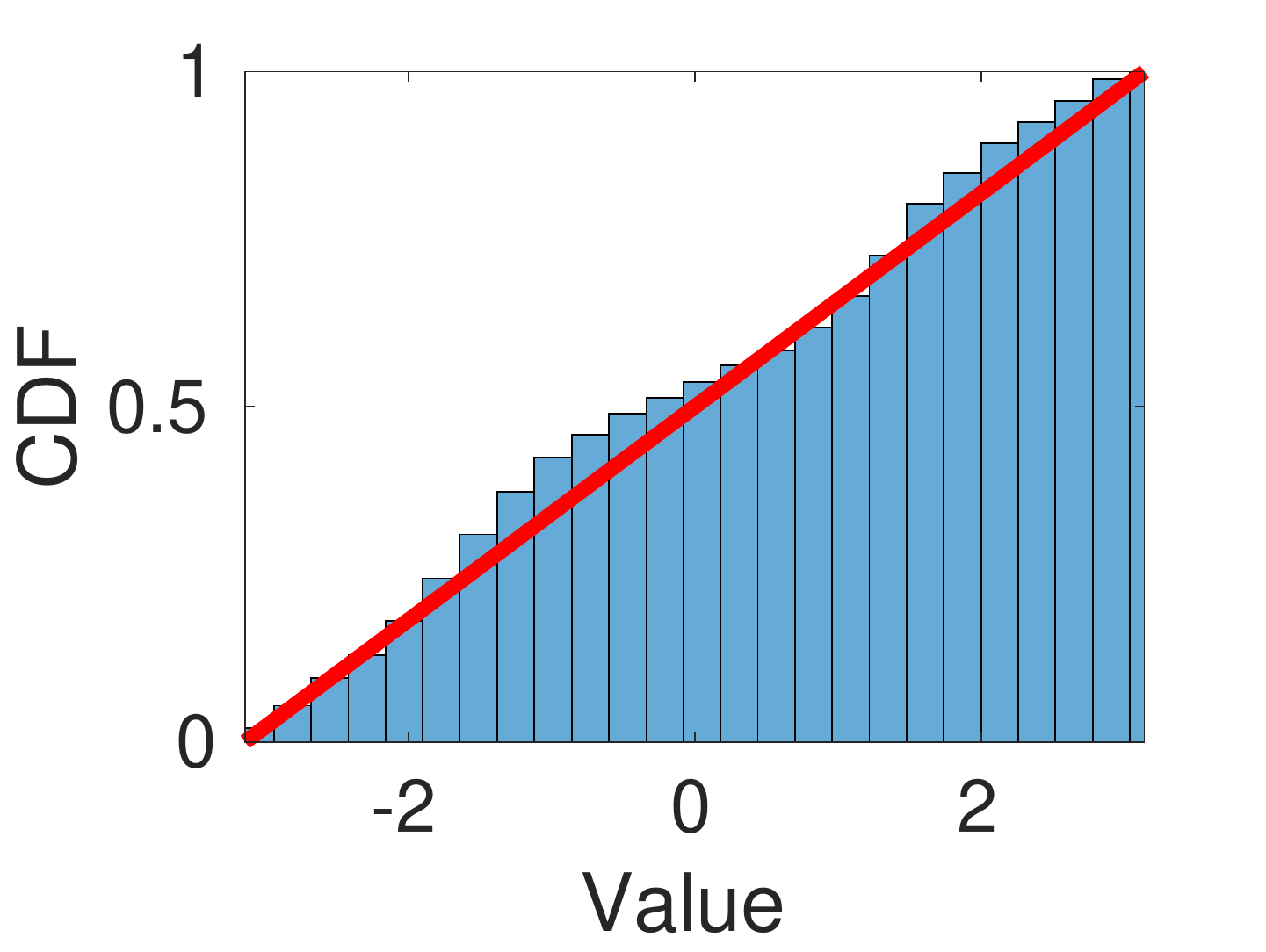}}
\newcommand{\addimbb}{\includegraphics[width=\curww\columnwidth]{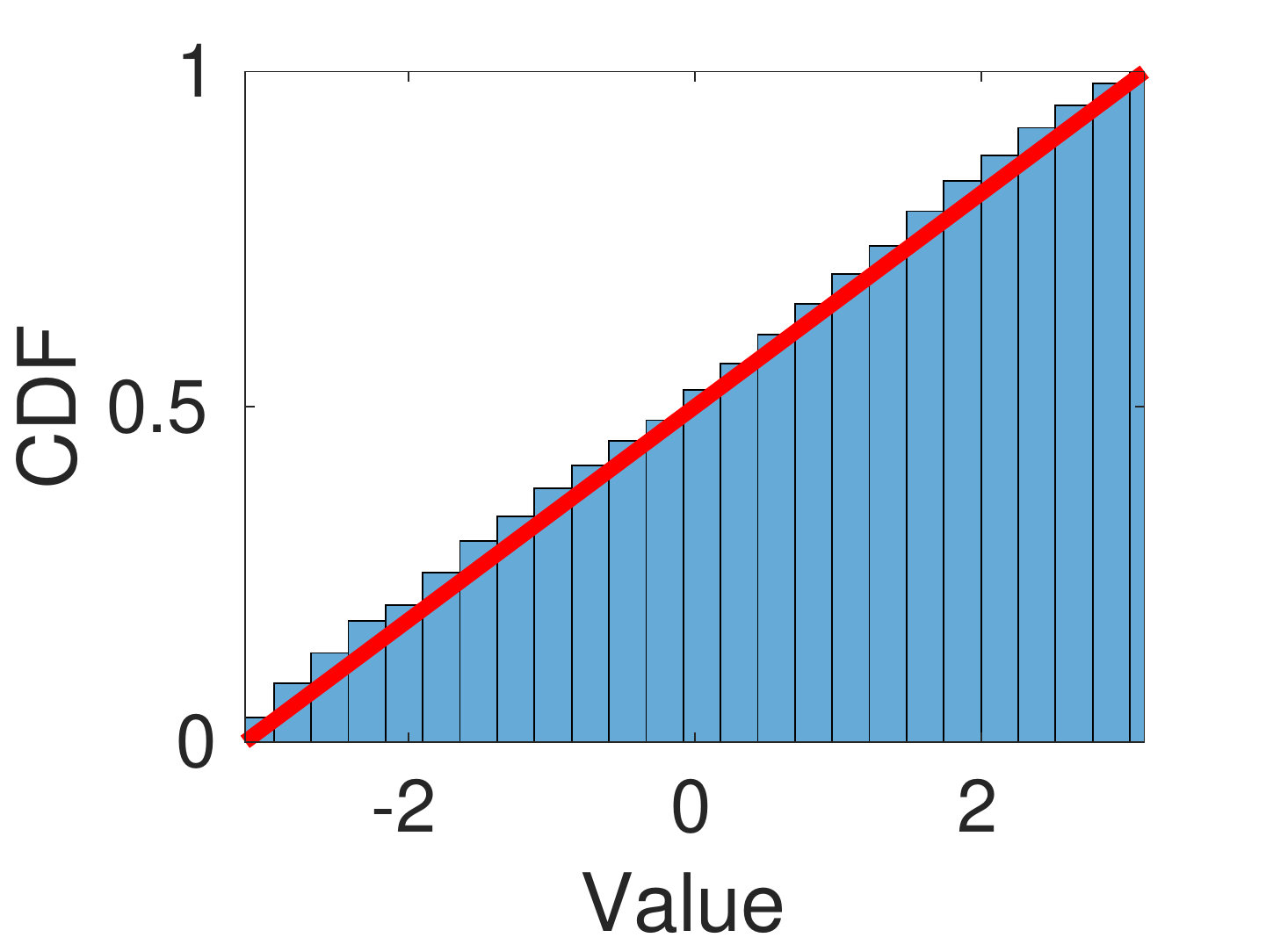}}
\newcommand{\addimbc}{\includegraphics[width=\curww\columnwidth]{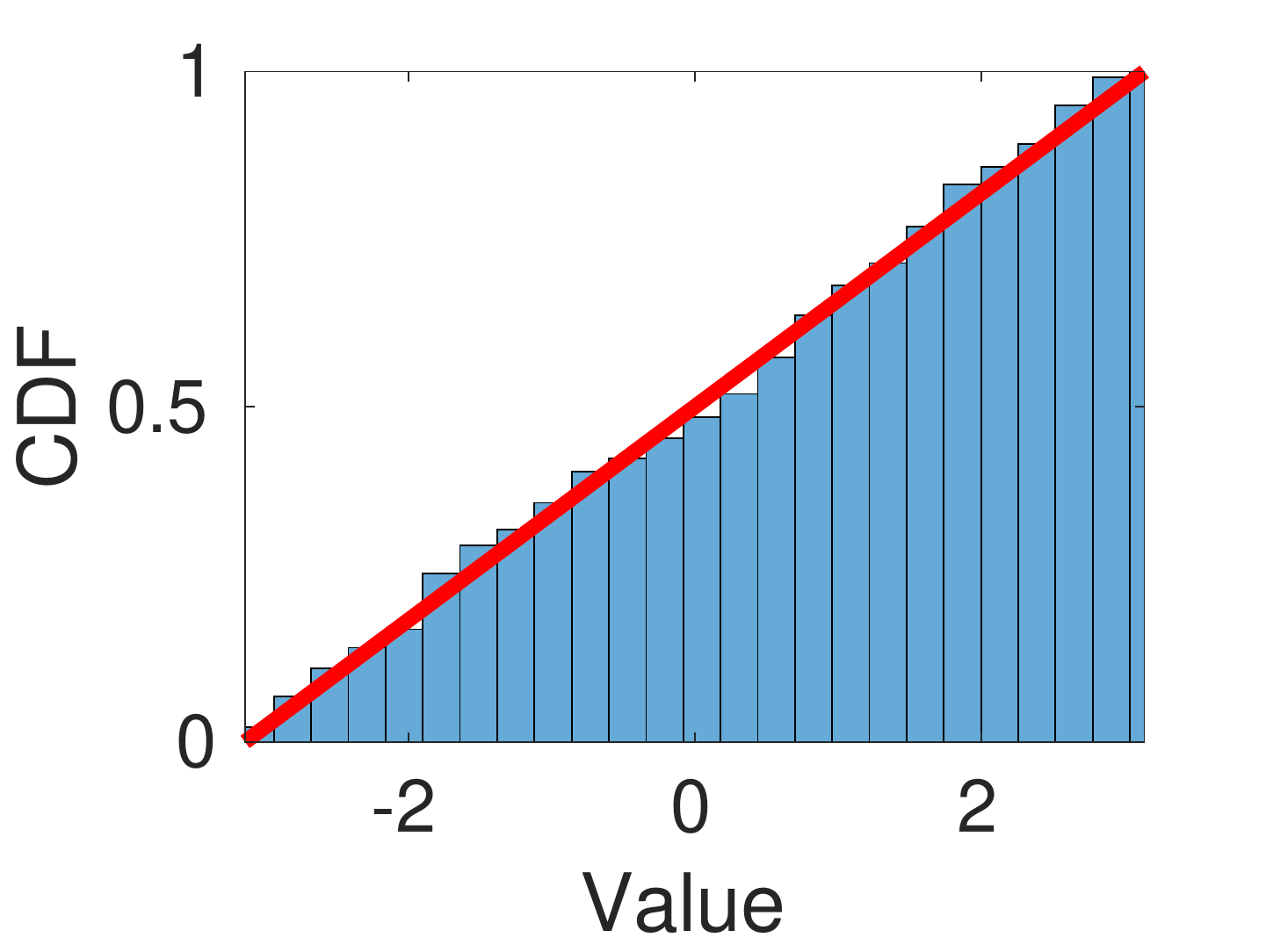}}
\newcommand{\imgat}{\protect 0.108}
\newcommand{\imgbt}{\protect 0.101}
\newcommand{\imgct}{\protect 0.193}
\newcommand{\imbat}{\protect 0.091}
\newcommand{\imbbt}{\protect 0.055}
\newcommand{\imbct}{\protect 0.043}

\renewcommand{\subs}{
\subfloat[Sc.2,{\bf S}, $D=\protect 0.108$. \label{figw:histsg2}]{\addimga} ~
\subfloat[Sc.3,{\bf S}, $D=\protect 0.101$. \label{figw:histsg3}]{\addimgb} ~
\subfloat[Sc.4,{\bf S}, $D=\protect 0.193$. \label{figw:histsg4}]{\addimgc} \\
\subfloat[Sc.2,{\bf W}, $D=\protect 0.091$. \label{figw:histsb2}]{\addimba} ~
\subfloat[Sc.3,{\bf W}, $D=\protect 0.055$. \label{figw:histsb3}]{\addimbb} ~
\subfloat[Sc.4,{\bf W}, $D=\protect 0.043$. \label{figw:histsb4}]{\addimbc} 
}
\begin{figure} \centering
\newcommand{\subreg}[1]{\protect\subref{figw:histsg#1}}
\newcommand{\subreb}[1]{\protect\subref{figw:histsb#1}}

\subs

\caption{Cdf of marginal histograms of wavelet coefficients of the first orientation of the "cameraman" image. Phase coefficients corresponding to high magnitudes (strong, denoted "{\bf S}") deviate from the uniform distribution, whereas phase coefficients corresponding to low magnitudes  (weak, denoted "{\bf W}") are of a distribution closer to uniform. Figs. \subreg{2},  \subreg{3} and \subreg{4}: Strong coefficients of scales 2, 3 and 4 (denoted by "Sc."), respectively. Figs. \subreb{2},  \subreb{3} and \subreb{4}: Weak coefficients of scales 2, 3 and 4 (denoted by "Sc."), respectively. The maximal difference between the cdf of the empirical and the uniform distribution are denoted by "D" on each figure.
\label{fig:Cdf-of-marginal}}
\end{figure}

\renewcommand{\curww}{0.45}
\renewcommand{\curo}{1}
\newcommand{\addimgh}{\includegraphics[width=\curww\columnwidth]{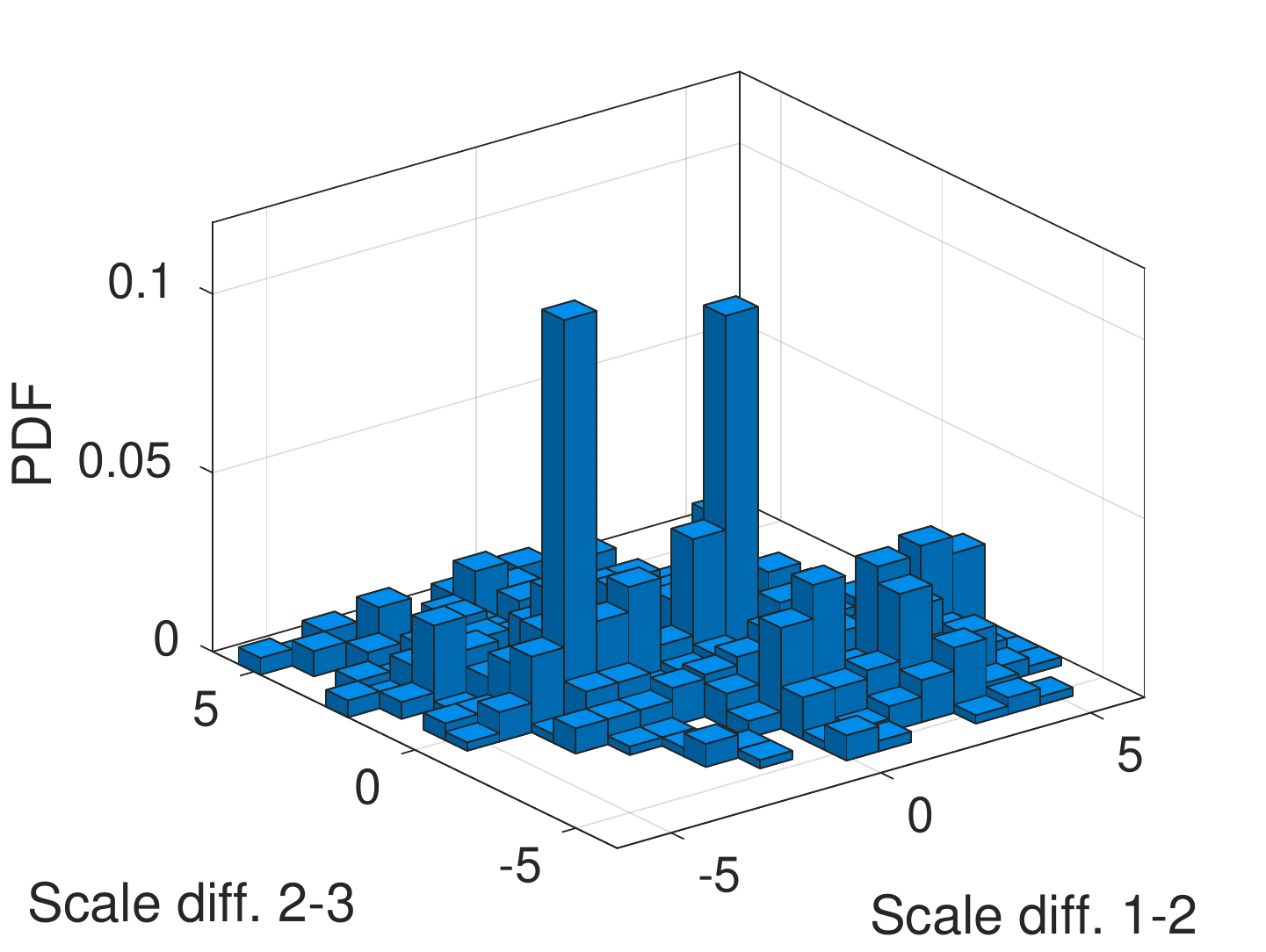}}
\newcommand{\addimbh}{\includegraphics[width=\curww\columnwidth]{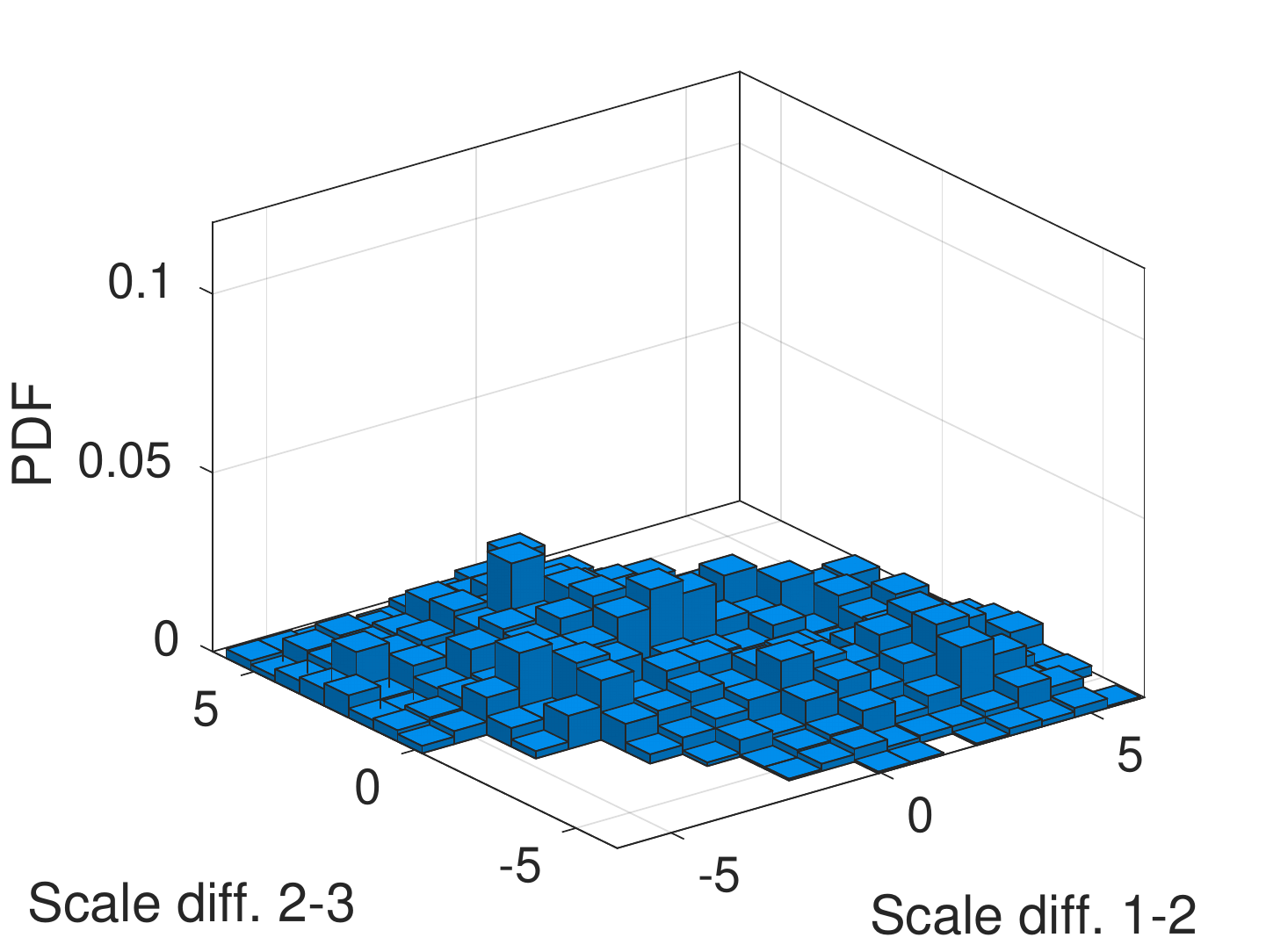}}
\newcommand{\imgh}{\protect 0.113}
\newcommand{\imbh}{\protect 0.030}

\renewcommand{\subs}{
\subfloat[Strong, $K=\protect 0.113$. \label{figw:hist2g}]{\addimgh} ~
\subfloat[Weak, $K=\protect 0.030$. \label{figw:hist2b}]{\addimbh}

}
\begin{figure}
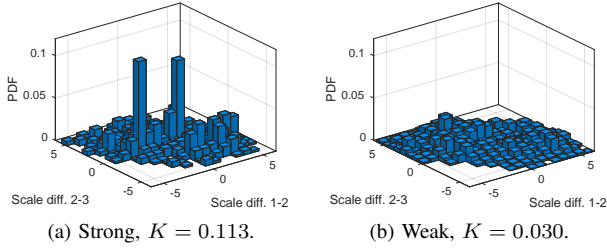
 \centering
\newcommand{\subreg}[1]{\protect\subref{figw:hist2#1}}

\subs
\caption{Pdf of adjacent scale joint histogram of wavelet phase coefficient difference of the "cameraman" image. Phase coefficients corresponding to high magnitudes ("Strong") are distributed differently than coefficients corresponding to low magnitudes  ("Weak"). Figs. \subreg{g} and  \subreg{b}: Strong and weak joint histograms of phase coefficient differences, respectively. The kurtosis of each empirical distribution are denoted by "K" on each figure.
\label{fig:Joint-histograms-of}}
\end{figure}

A similar phenomenon is observed by inspecting 2D histograms, revealing dependency. We calculate the joint empirical pdf of the \emph{phase difference} between the second-third and third-fourth scales (Fig. \ref{fig:Joint-histograms-of}), and compare the maximal count frequency (most occurring combination) between high- and low-energy coefficients, defined by thresholding the magnitudes.

To observe this phenomenon better, we analyze the coefficients over a range of thresholds, for $20\%$ to $7\%$ of the strongest coefficients (this range captures the differences between the high- and low-energy coefficients), and observe that the maximal count frequency for high-energy coefficients is considerably higher in almost all cases, indicating that this phenomenon does not depend considerably on the threshold (Fig. \ref{fig:Maximal-frequency-for}).

The high maximal frequency in the examples (Fig. \ref{fig:Maximal-frequency-for} and Fig. \ref{fig:Joint-histograms-of}) corresponds to a peak in the probability distribution of the coefficients. This indicates that for high-energy coefficients, there is lower variance in the probable values than for low-energy coefficients, corresponding to image structure encoded in the former.

\renewcommand{\curww}{0.33}
\newcommand{\addimaxo}[1]{\includegraphics[width=\curww\columnwidth,clip=true,trim=5 0 30 0]{coeff_visu/im1_max_o_#1_review1}}

\renewcommand{\subs}{
\subfloat[Orient. 1 \label{figw:histma}]{\addimaxo{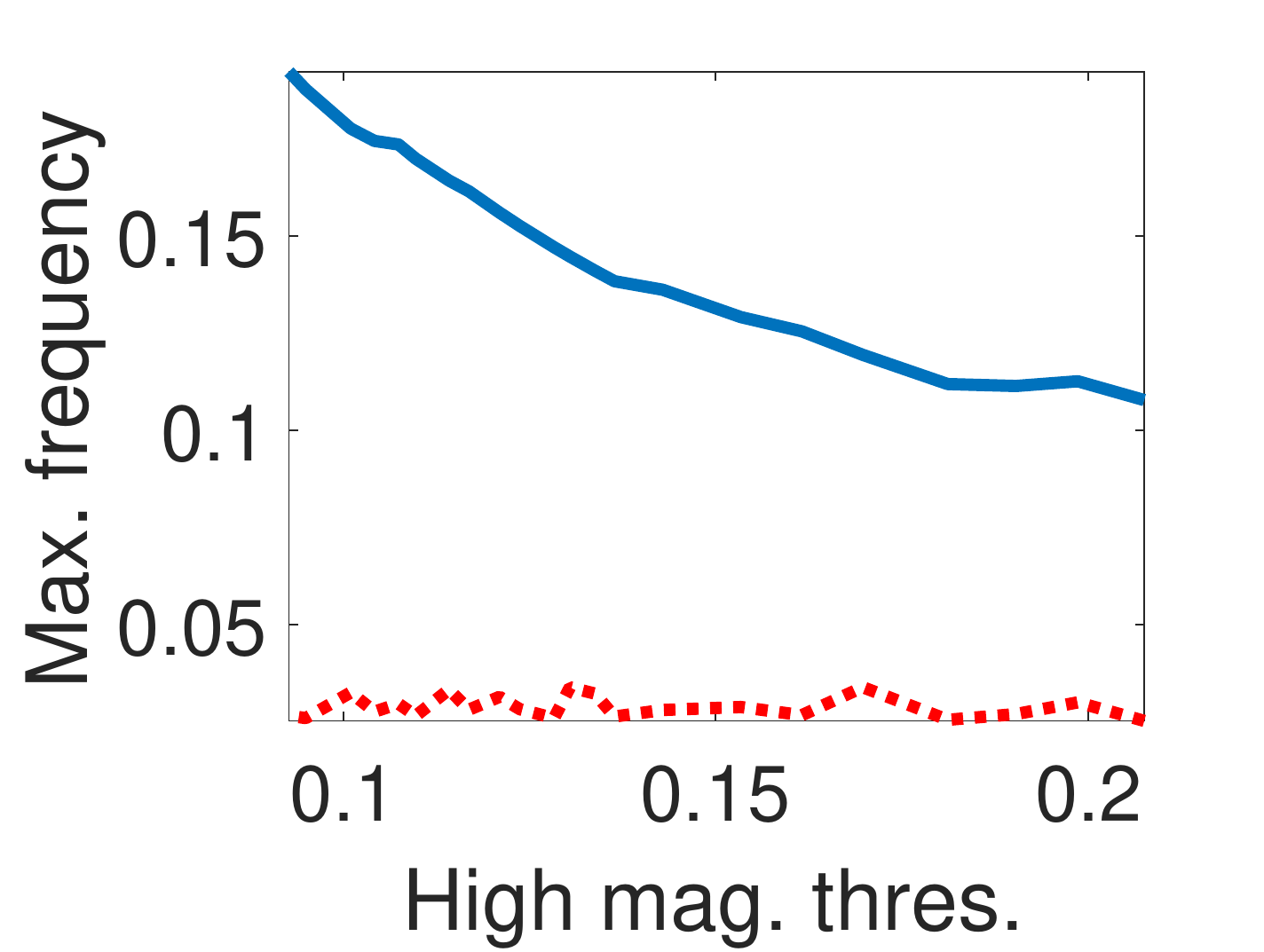}} 
\subfloat[Orient. 2 \label{figw:histmb}]{\addimaxo{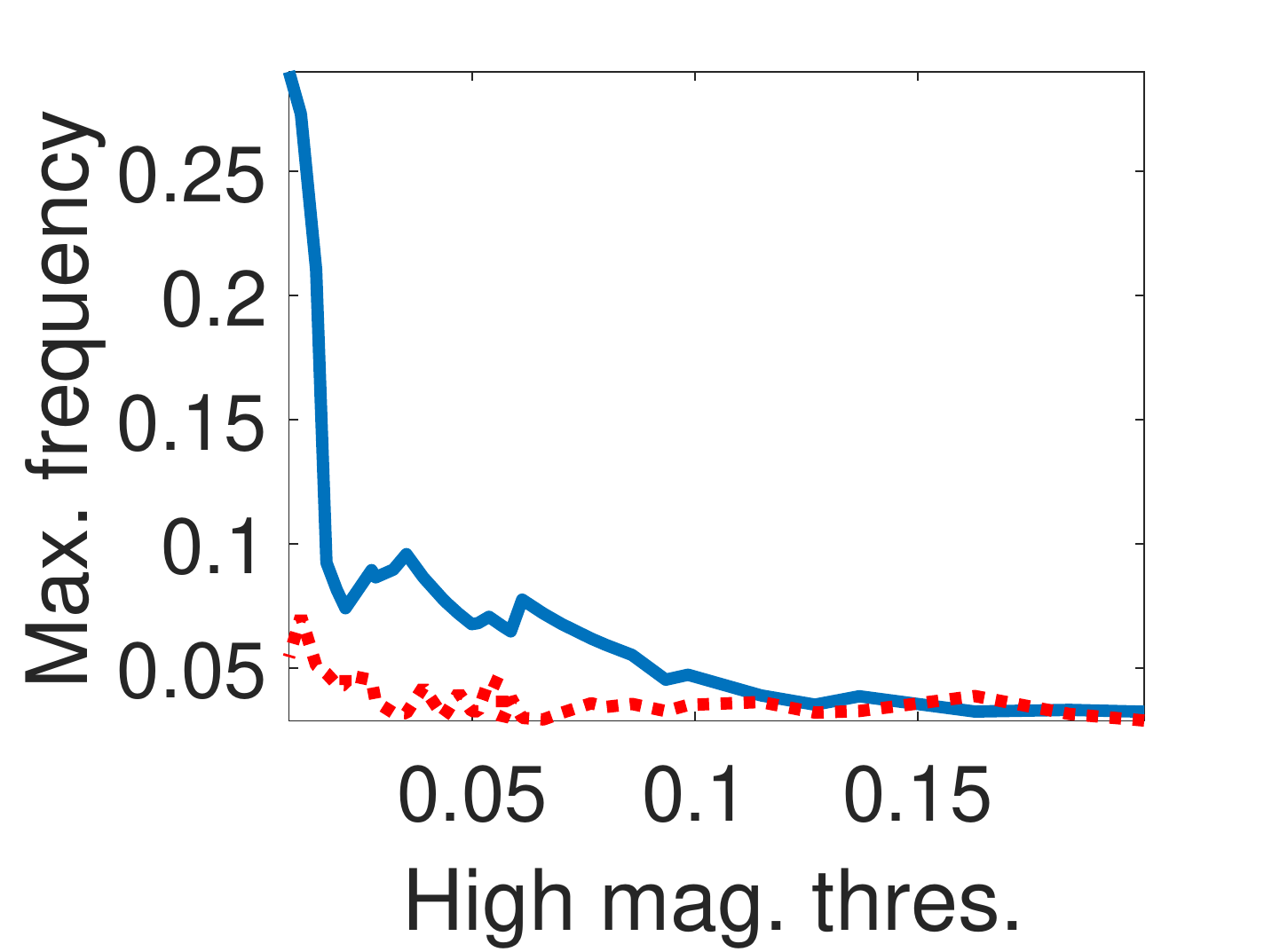}} 
\subfloat[Orient. 3 \label{figw:histmc}]{\addimaxo{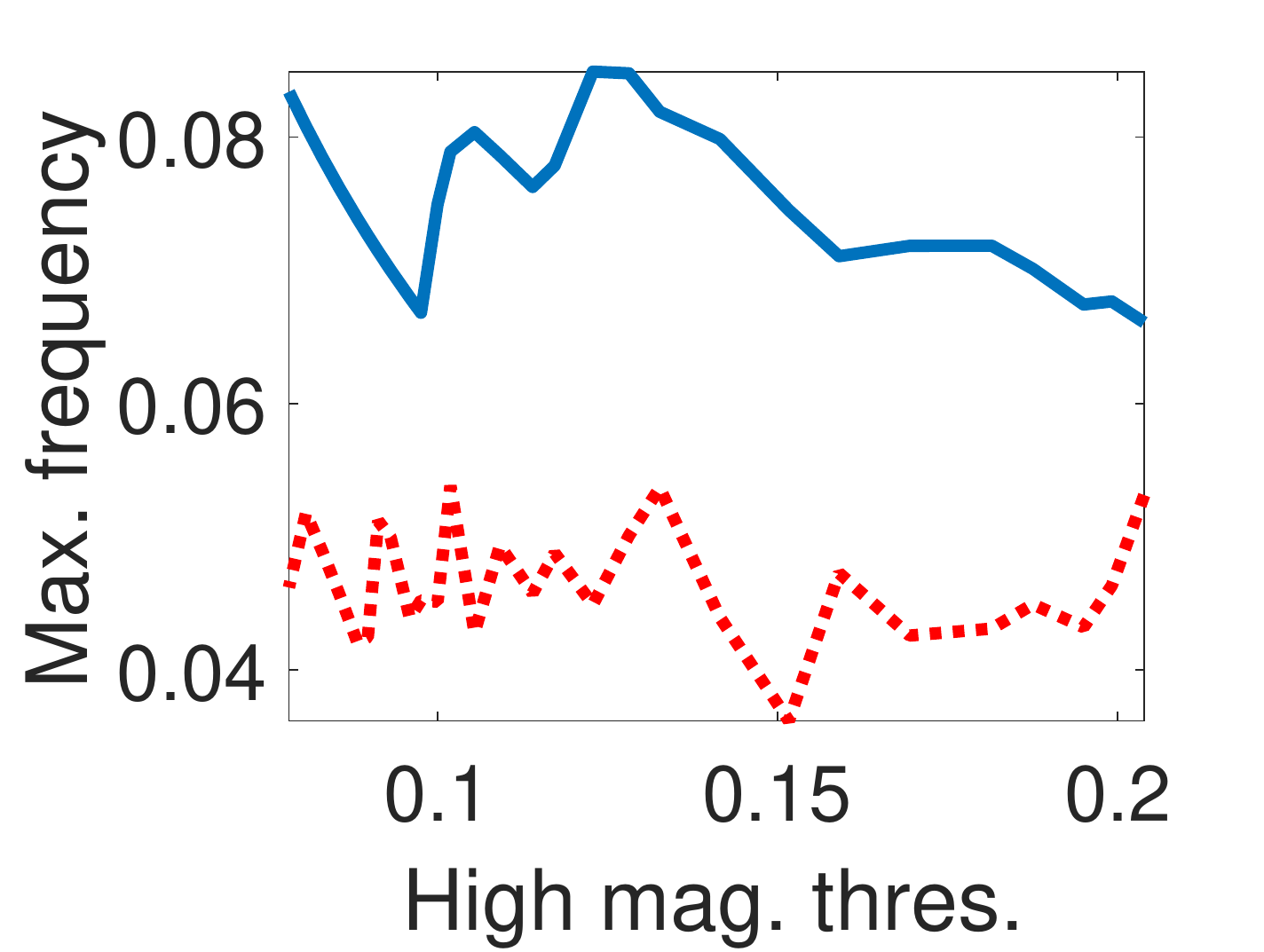}} \\
\subfloat[Orient. 4 \label{figw:histmd}]{\addimaxo{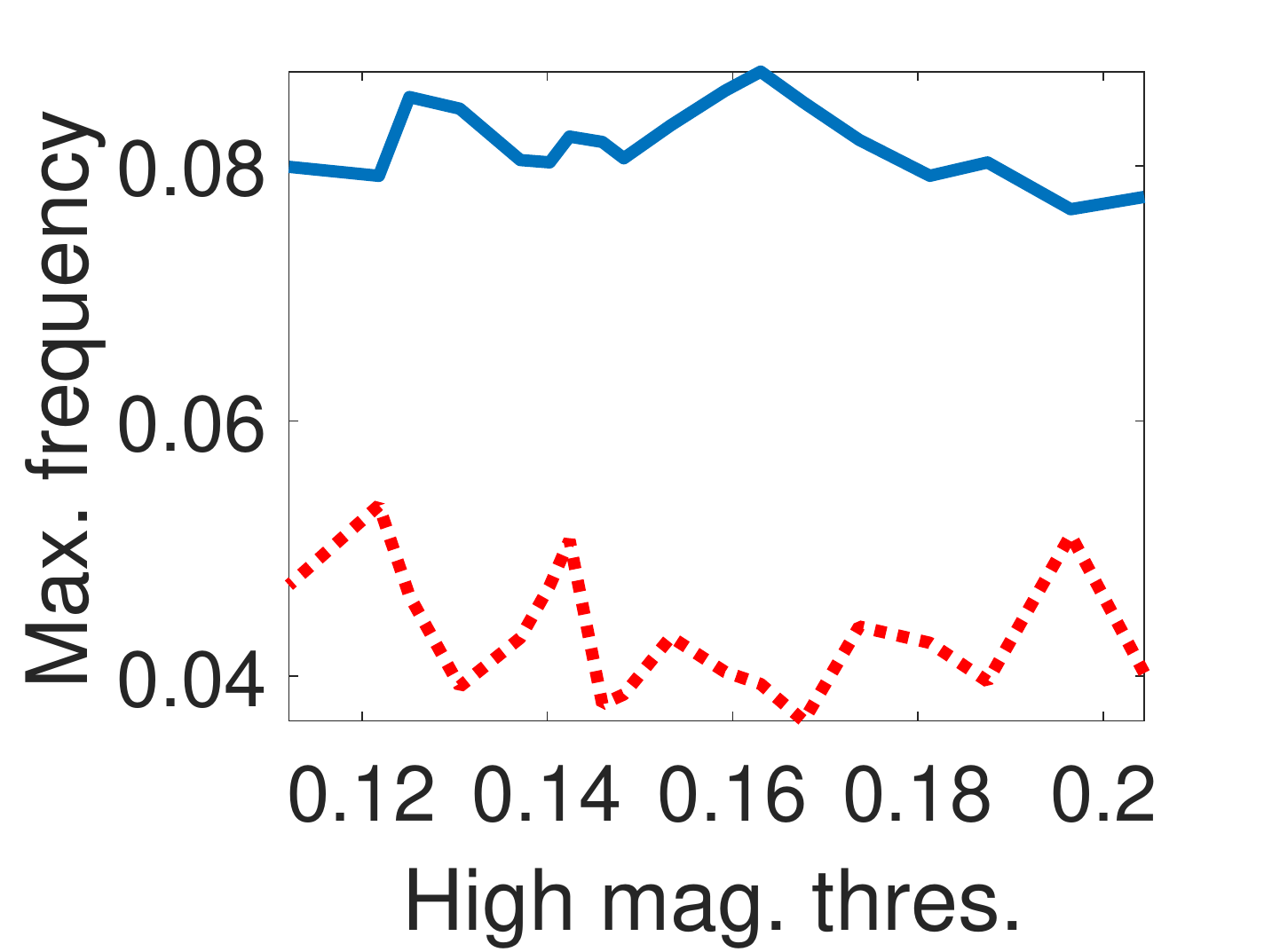}} 
\subfloat[Orient. 5 \label{figw:histme}]{\addimaxo{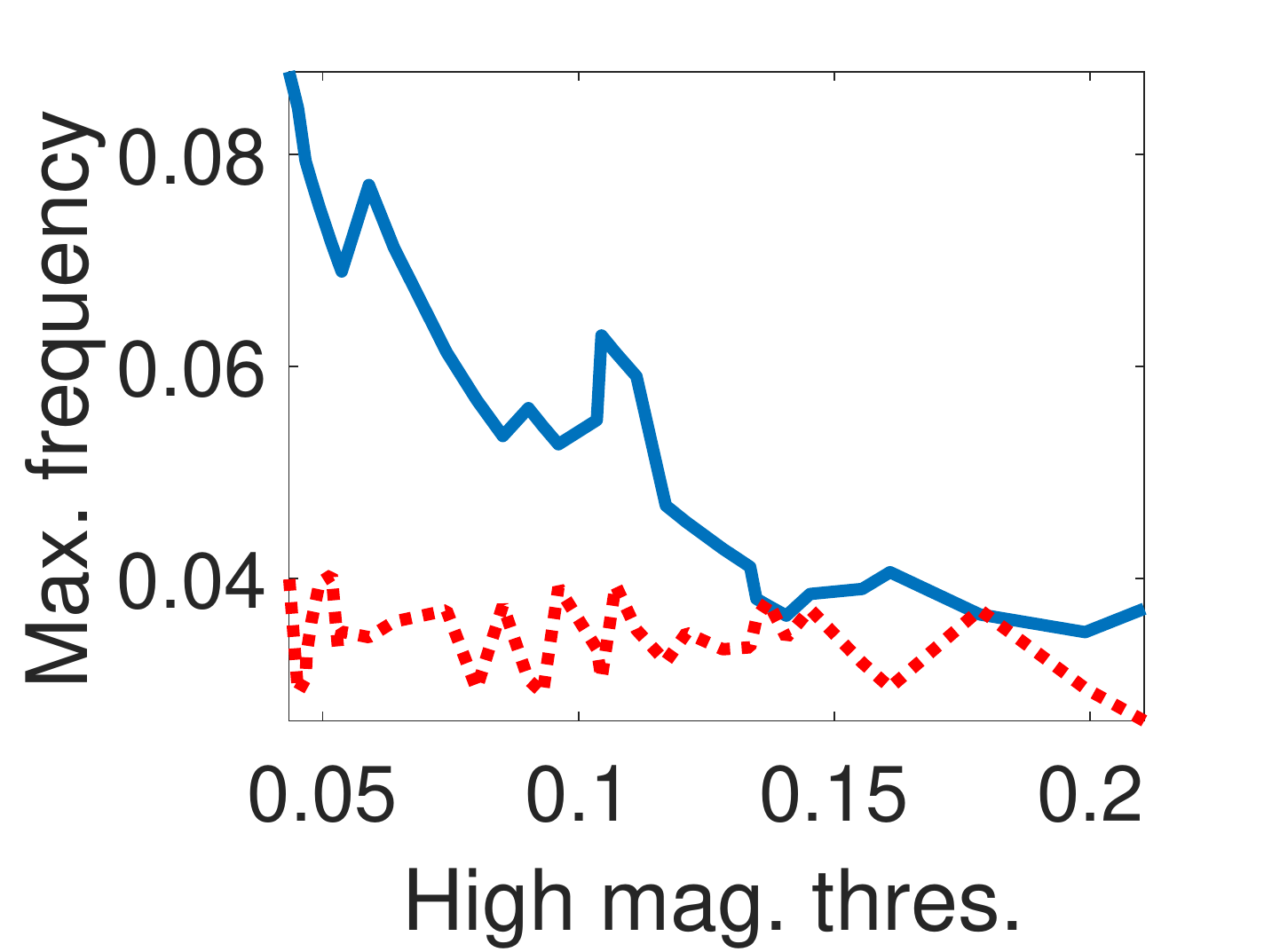}} 
\subfloat[Orient. 6 \label{figw:histmf}]{\addimaxo{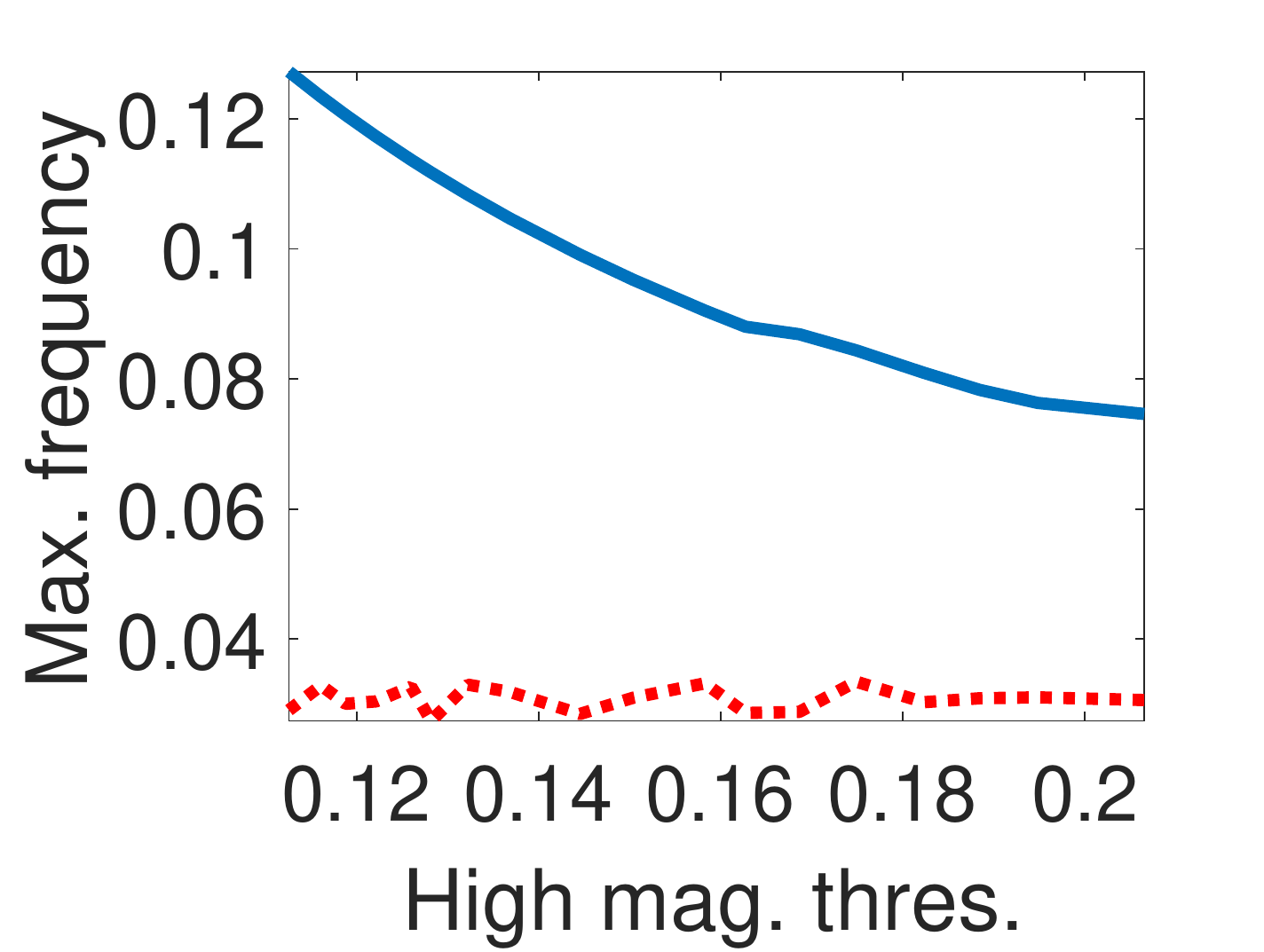}}
}

\begin{figure} \centering
\newcommand{\subreg}[1]{\protect\subref{figw:histm#1}}

\subs
\caption{Maximal count frequency for phase difference joint histograms. Phase coefficients corresponding to high magnitudes have much higher maximal frequencies (denoted by solid blue lines), revealing less random behavior, compared with low-energy coefficients (denotes by dashed red lines).  \subreg{a}--\subreg{f} show maximal frequencies of orientations 1--6, respectively (denoted "Orient."), for varying thresholds of high vs. low magnitudes.
\label{fig:Maximal-frequency-for}}
\end{figure}

\section{A Gaussian mixture model for local phase }

\global\long\def\one{\mathbbm{1}}

Due to statistical dependencies between adjacent scales and spatial
locations, observed in this work and elsewhere, we model a phase coefficient
along with its immediate neighborhood, as expressed in the local graph
excerpt (Fig. \ref{fig:A-local-graph}). We thereby have an underlying
Markov assumption, that phase coefficients can be described by means
of their local neighbors. This modelling allows us to estimate distributions
based on moderately-sized image banks, due to the fact that the dimension
of the random vector is $10$ (Section \ref{sec:decomposition-graph}). Following our discussion as to the
importance of thresholding high-energy coefficients, our model is
based on high-energy coefficients only, where the threshold is a parameter
that needs to be imposed. In our experiments we set the threshold
so that $20\%$ of highest magnitudes are considered.

We define the sub-tree as an excerpt of the complete wavelet graph
tree (Fig. \ref{fig:A-local-graph}); it is vectorized to a $10$-dimensional
vector, $\boldsymbol{v},$ as follows: $v\left(1\right)$ is the middle
node, $v\left(2\right),...,v\left(5\right)$ are the adjacent nodes
in the same level, $v\left(6\right),...,v\left(9\right)$ are the
child nodes, and $v\left(10\right)$ is the parent node. Using the notations
of Fig. \ref{fig:A-local-graph}, we have 
\begin{align}
v= & \left[c_{x,y}^{i},c_{x-1,y}^{i},c_{x,y-1}^{i},c_{x+1,y}^{i},c_{x,y+1}^{i}\right.,\nonumber \\
 & \left.c_{x-1,y}^{i+1},c_{x,y-1}^{i+1},c_{x+1,y}^{i+1},c_{x,y+1}^{i+1},c_{x,y}^{i-1}\right].\label{eq:subtree_vec}
\end{align}

The sub-trees considered for learning are only ``full'' neighbors that
belong to the detail wavelet coefficient. We do not model the approximation
(scaling) decomposition level. We use $384$ images, randomly selected
from the Brodatz and McGill texture datasets, cropped to size $N\times N$,
where $N=256$ with $4$ wavelet decomposition levels. From each image
we extract the $N^{2}/64$ coefficients with the highest magnitudes
and their respective sub-trees, which yields the approximate $20\%$
threshold.

We train the GMM via the expectation-maximization (EM) algorithm.
The number of parameters to be learned is defined as:
\[
n_{params}=\left(m^{r}+m\right)\cdot k+k,
\]
where $m$ is the vector's dimension, $k$ is the number of components,
and the value of $r$ is either $1$ or $2$, depending on the covariance
structure, where $r=1$ for diagonal covariance and $r=2$ for full
covariance. In our case, $m=10$, $k=10$ and $r=1$. The reasoning
for selecting $k=10$ is explained in the sequel. We define the following
ratio:
\[
r_{params}=n_{samples}/n_{params},
\]
where $n_{samples}$ is the number of samples available for learning.
To provide sufficient samples for EM-based learning, we demand $r_{params}>50$,
so that for each parameter learned there are at least $50$ samples. 

Despite the fact that we observe dependencies in joint histograms
of phase differences (e.g. Fig. \ref{fig:Joint-histograms-of}), we learn the phases values without applying
any difference or other linear schemes, under the assumption that
the EM algorithm will learn the underlying structure of the phase,
provided there are sufficient samples.

In the sequel we demonstrate the properties of the learned phase model
as well as use it for various applications in image processing.

\subsection{Choosing the right number of components}

Gaussian mixtures can effectively describe any distribution, provided
the number of components is high enough. However, using too many Gaussian
components can lead to over-fitting and loss of generality. We use
K-fold cross validation to choose an optimal number of components.
Given a set of $M$ sub-tree vectors, we partition the dataset to
$U=10$ randomly selected equal subsets. We then train the GMM for
$K$ components using all but one subset, and calculate the mean of
the log-likelihood of each of the left-out variables. This is iterated
$U$ times, where at every iteration, a different subset is left out.
The mean log-likelihood is then averaged again across all iterations.
The result provides a scalar number that indicates how good is the definition 
of the test data (the left-out subset) according to the model trained by the data. 

\begin{figure}
\noindent \begin{centering}
\includegraphics[width=0.8\columnwidth]{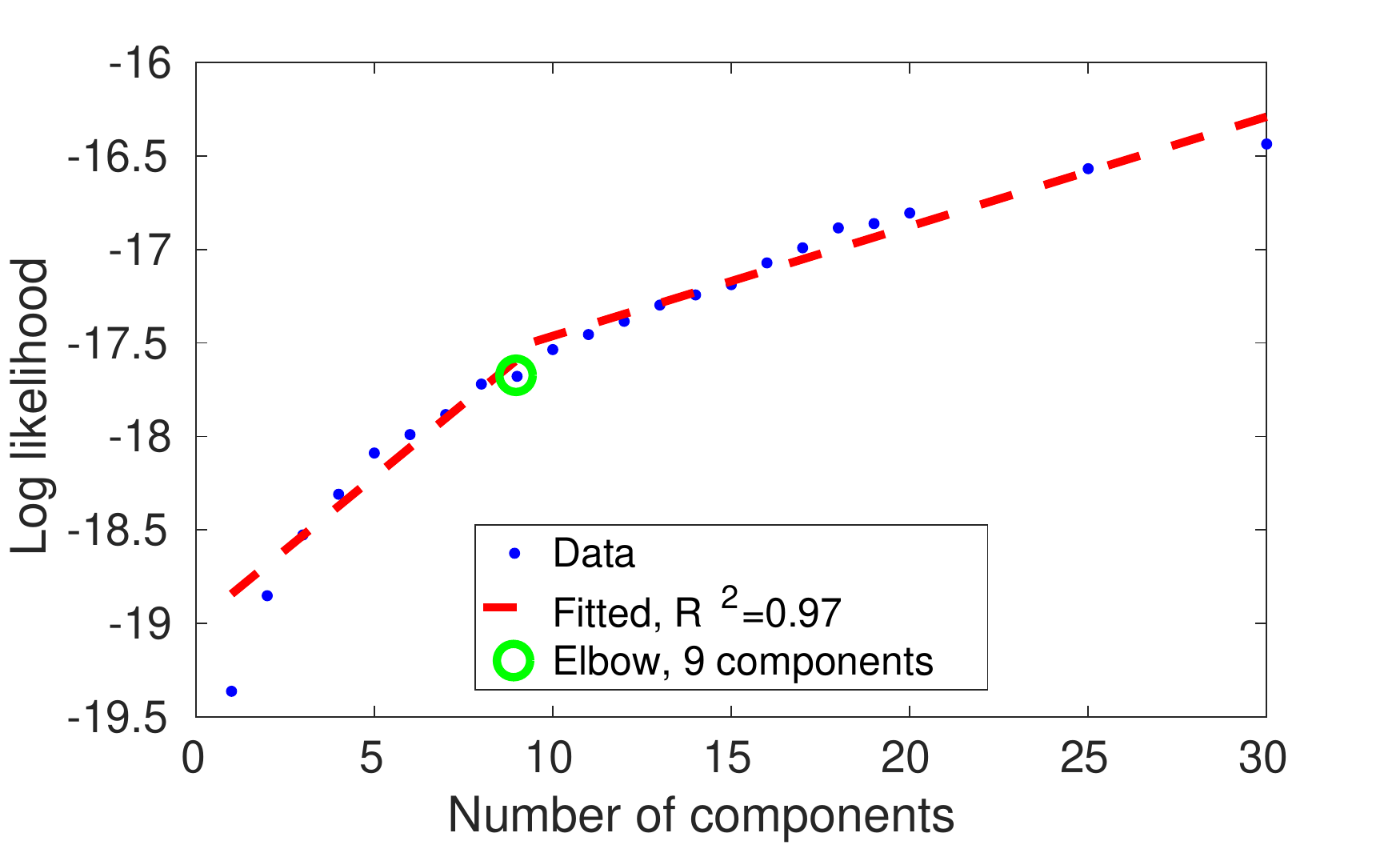}
\par\end{centering}
\caption{Estimating the number of components for modelling local-phase with
GMM. The elbow was found to be between $5$ and $10$ components.
\label{fig:Estimating-the-number}}

\end{figure}

The optimal mixture number is selected by finding the so-called \emph{elbow}
in the mean log-likelihood, plotted against the number of components.
More components will describe the test data better, due to more dense
filling of the feature space and will therefore not decrease, but
the contribution of more components after a certain threshold will
be marginal. This is indicated by an elbow in the mean log-likelihood
vs. number of component, and this is the chosen number of mixture
components we use \cite[§14]{Hastie2009}. We observe that a number
of $5$ to $10$ components, as used in this work, fits the elbow
assumption (Fig. \ref{fig:Estimating-the-number}). The elbow was
found automatically by fitting two first order polynomials to the
log-likelihood of the test set, when the GMM has been fitted based
on the training data with a set number of components. We note that the use of $10$ components instead of $9$ (Fig. \ref{fig:Estimating-the-number}) does not affect the model in any noticeable manner.

Since in all our experiments (including those that are not shown here) we use up to $30$ components, a large amount of data is required for training. The aforementioned number of images
($384$) yields a ratio of samples to trainable parameters of at
least $600$ and the number of sub-tree vectors used was $M\approx393\times10^{3}$, verifying that sufficient data was available.

\subsection{Demonstration of the phase model }

The learned model yields $K=10$ normally distributed components.
In this section, we demonstrate intimate relationship between the learned
coefficients and the visual aspects that relate to coherent structures.
We define the average congruency ($AG$), similarly to the phase congruency \cite{Kovesi1999a}, as follows:
\[
AG\left(k\right)=\left|\frac{1}{M}\sum_{m=1}^{M}\cos\left(\eta_{k}\left(m\right)-\bar{\eta}_{k}\right)\right|,
\]
where $M$ is the number of scales considered in the model (in our
case $M=3$ since we analyze a tree that spans at most three scales),
where lower values of $m$ correspond to coarser scales; $k$ is the
component number, $\eta_{k}\left(m\right)$ is the mean value of the
phase at the $m$th scale in the $k$th component, as learned via
the Gaussian model. $\bar{\eta}_{k}$ is the mean value of $\eta_{k}$.
Using the sub-tree structure (\ref{eq:subtree_vec}) we define:
\begin{align*}
\eta_{k}\left(1\right) & =2^{1}\cdot\mu_{k}\left(10\right)\\
\eta_{k}\left(2\right) & =2^{0}\cdot\mu_{k}\left(1\right)\\
\eta_{k}\left(3\right) & =2^{-1}\cdot\left[\frac{1}{4}\sum_{i=6}^{9}\mu_{k}\left(i\right)\right].
\end{align*}
This is explained as follows: the first mean phase coefficient is
the coarser phase, $\mu_{k}\left(10\right)$. The next phase coefficient
is the central phase coefficient, $\mu_{k}\left(1\right)$, and the
finest phase coefficient is a mean of the four children phase nodes,
$\mu_{k}\left(i\right)$ for $i=6,...,9$. The phases are normalized
by powers of $2$ according to their relative scales to compensate
for the dyadic decimation that affects the phase, as was described
elsewhere \cite{Miller2006,Miller2008}. 

The average congruency, $AG\left(k\right)$, reflects the phase congruency
quantity \cite{Kovesi1999} for cases in which the magnitude is equal
in all coefficients. The value of $AG\left(k\right)$ reflects the
same meaning as the original phase congruency, as low values (close
to $0$) correspond to incoherent structures with phases in random
directions, whereas high values (close to $1$) correspond to coherent
structures in which all the cosines are close to $1$.

We use this quantity to assess the coherence of each component in
the Gaussian mixture; we expect the model to learn components that
correspond to high $AG\left(k\right)$ values and represent coherent
structures, as well as components that correspond to low $AG\left(k\right)$
values and represent incoherent structures.

To demonstrate this effect, we train the model using $20$
images from the Brodatz dataset. We then decompose an image with coherent
structures to its wavelet graph (Fig. \ref{visuxbox}), and for each
pixel in the intermediate scales we calculate the posterior component,
$\P{k|y}$. Given a component number, $k$, we place markers on the
decomposition images if the coefficient was derived from the $k$th
component, defined as the component that maximizes the posterior probability,
provided the probability is higher than a threshold set to $0.8$.

Using this demonstration scheme, we plot the markers on the decomposition
images for several values of $k$, which correspond to components
with different values of $AG\left(k\right)$. We observe that, indeed,
when we choose $k$ with high $AG\left(k\right)$, the markers are
placed on edge-type structures, and vice versa; low values of $AG\left(k\right)$
correspond to coefficients that contain mostly noise or other non-coherent
structure. 

\newcommand{\visu}[3]{\begin{figure}
\subfloat[$S2,AG\,0.84$]{ \includegraphics[width=#3\columnwidth,clip=true,trim=80 150 80 150]{#1sc21} }
\subfloat[$S2,AG\,0.78$]{ \includegraphics[width=#3\columnwidth,clip=true,trim=80 150 80 150]{#1sc22} }
\subfloat[$S2,AG\,0.34$]{ \includegraphics[width=#3\columnwidth,clip=true,trim=80 150 80 150]{#1sc24} }
\subfloat[$S2,AG\,0.01$]{ \includegraphics[width=#3\columnwidth,clip=true,trim=80 150 80 150]{#1sc25} }
\\
\subfloat[$S3,AG\,0.84$]{ \includegraphics[width=#3\columnwidth,clip=true,trim=80 150 80 150]{#1sc31} }
\subfloat[$S3,AG\,0.78$]{ \includegraphics[width=#3\columnwidth,clip=true,trim=80 150 80 150]{#1sc32} }
\subfloat[$S3,AG\,0.34$]{ \includegraphics[width=#3\columnwidth,clip=true,trim=80 150 80 150]{#1sc34} }
\subfloat[$S3,AG\,0.01$]{ \includegraphics[width=#3\columnwidth,clip=true,trim=80 150 80 150]{#1sc35} }
\caption{#2}
\end{figure} }

\visu{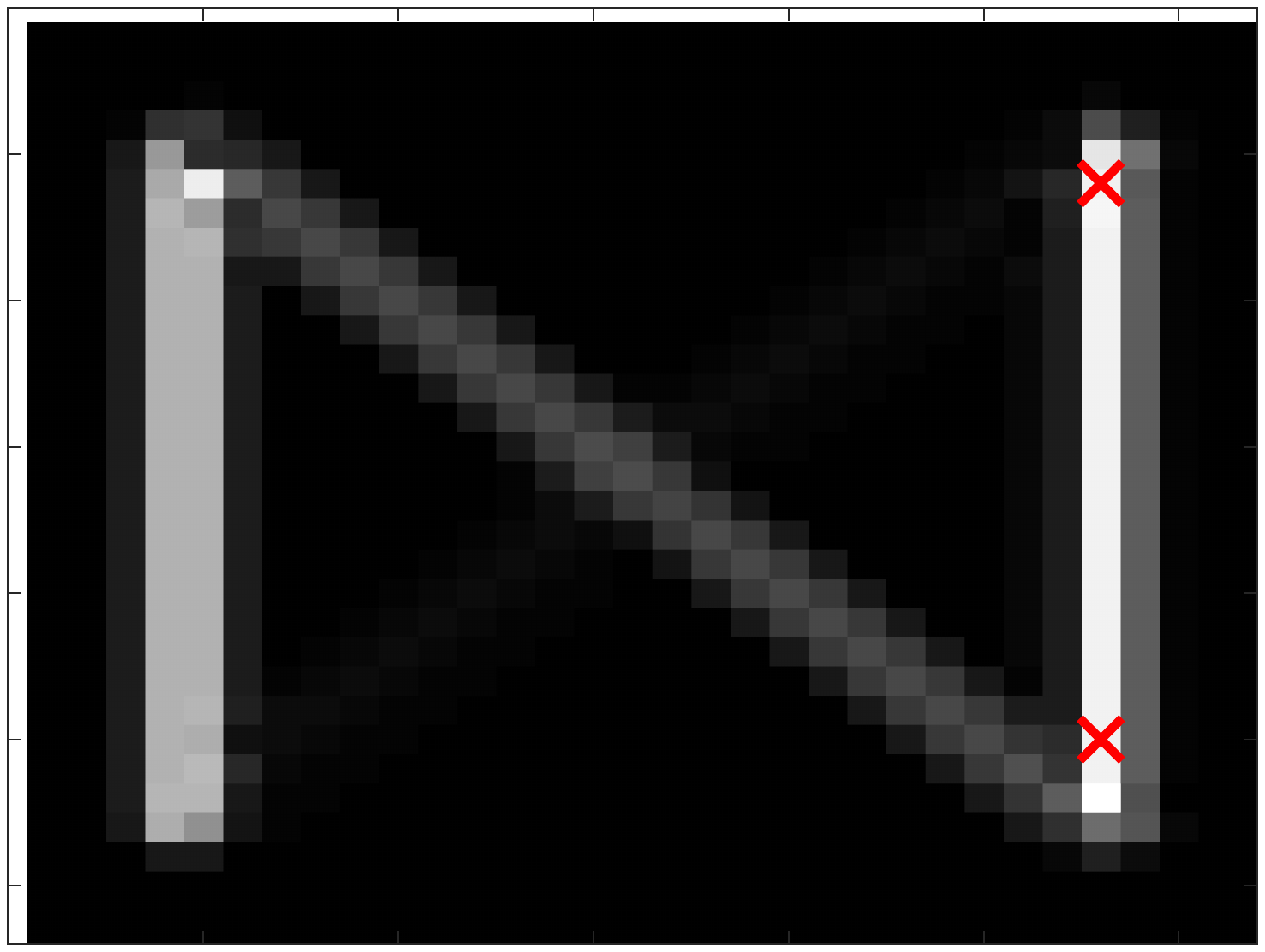}{Demonstration of the local phase model using a synthetic image. Shown is the third orientation. First and second rows depict the second and third wavelet scales, respectively. The columns correspond to components in decreasing values of average congruency; in the first two columns we depict the two highest values and in the last columns depicted are the lowest two values. The red markers highlight coefficients that maximize the posterior for the given component. We observe that components with high values of average congruency are the posterior components for coherent local structures, such as edges, whereas less order and coherence is observed for structures with posterior components of low average congruency.\label{visuxbox}}{0.22} 

\begin{figure}
\noindent \begin{centering}
\includegraphics[width=0.5\columnwidth]{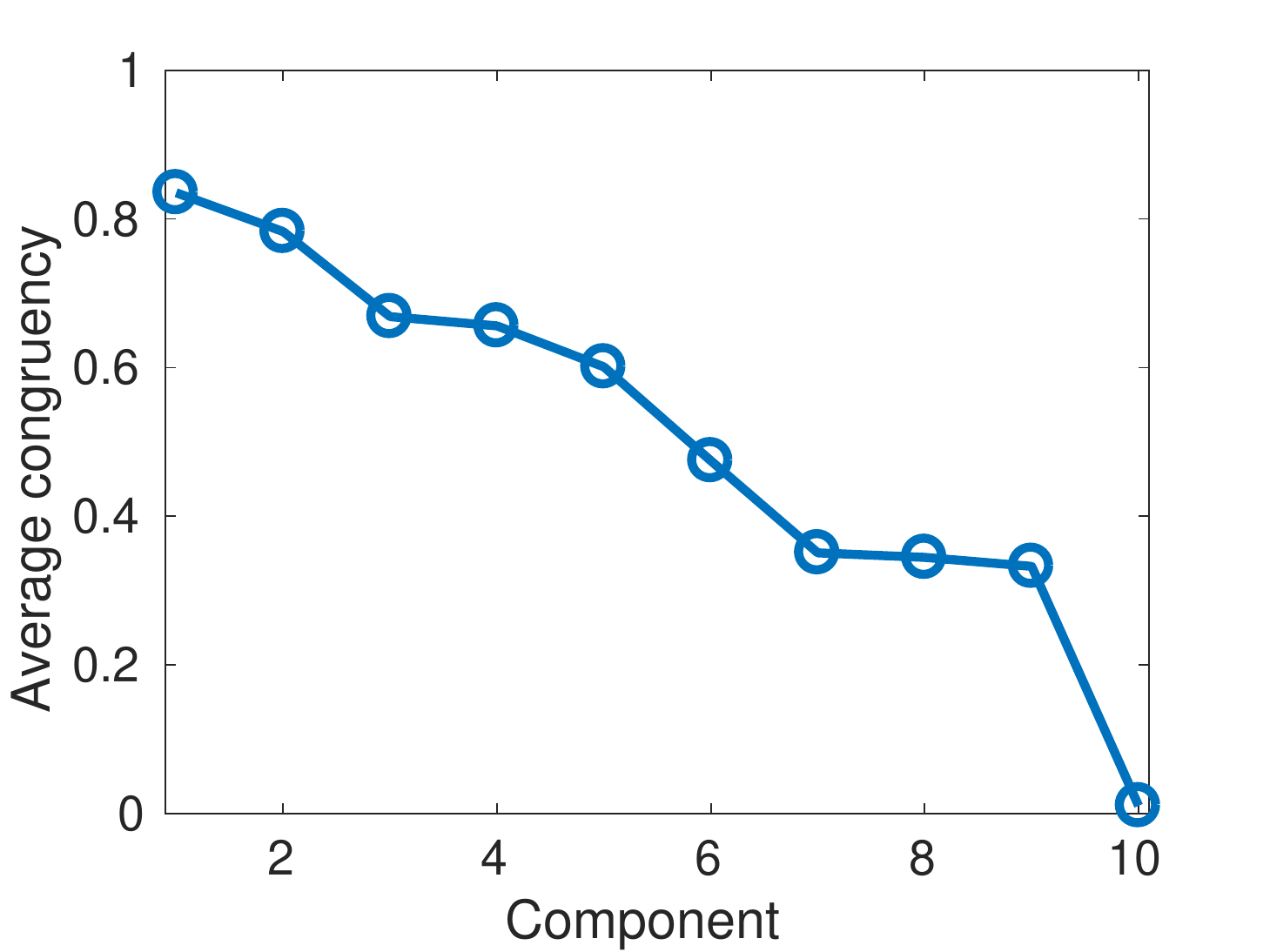}
\par\end{centering}
\caption{The average congruency as captured by each component of the model
(in descending order). The model captures various degrees of congruency.
\label{fig:The-average-congruency}}
\end{figure}
Further, we expect the model to capture various degrees of coherence,
varying from very coherent to non-coherent coefficients, as observed
in the learned model (Fig. \ref{fig:The-average-congruency}).

\section{Phase estimation under noisy conditions}

The first application we present is useful when the local phase is
noisy, due to distortions or lossy coding of the coefficients. We
use this scheme in order to demonstrate the efficacy of the proposed
model; we therefore show several examples of noisy phase which distorts
the image structure, and the reconstructed versions using the proposed
model.

Let $\eta\sim\mathcal{G}$ denote the detail phase coefficients and
the learned GMM distribution, $\mathcal{G}$. Let $\theta$ denote
the noisy phase:
\[
\theta=\eta+w,
\]

where $w$ is white Gaussian noise with variance $\sigma^{2}$. We
estimate $\eta$ given $\theta$:
\begin{align*}
\hat{\eta}\left(\theta\right) & =\E{\eta|\theta}=\int xp_{\eta|\theta}dx\\
 & =\int x\frac{p_{\theta|\eta}p_{\eta}}{p_{\theta}}dx=\frac{1}{p_{\theta}}\int xp_{\theta|\eta}p_{\eta}dx.
\end{align*}
The distribution of the noisy coefficients, $p_{\theta}$, is also
a GMM \cite{Cao2008} with the same parameters other than the covariance
which satisfies $\Sigma_{k}'=\Sigma_{k}+\sigma^{2}I$, where $\Sigma_{k}'$
and $\Sigma_{k}$ are the $k$'th covariances for $\theta$ and $\eta$,
respectively. This also provides means to estimate the GMM component
covariances given the noisy patch. 

$p_{\eta}$ is the probability for the learned GMM. $p_{\theta|\eta}$
is the noise distribution with parameter $\theta-\eta$: 
\[
q\left(\theta-\eta\right)\triangleq p_{\theta|\eta}\left(\theta,\eta\right)=\mathcal{N}\left(\theta-\eta;0,\sigma^{2}I\right).
\]

The joint distribution is 
\[
p_{\theta|\eta}p_{\eta}=\sum\pi_{k}g_{\eta}^{k}\cdot q\left(\theta-\eta\right),
\]

and the final estimation is given by \cite{Cao2008}:
\begin{align}
\hat{\eta}\left(\theta\right) & =\frac{1}{p_{\theta}}\sum_{k=1}^{K}\pi_{k}g_{\theta}^{k}\cdot\left(\Sigma_{k}'\right)^{-1}\left(\sigma^{2}\mu_{k}+\Sigma_{k}\theta\right).\label{eq:phase_denoising}
\end{align}

We note that this can be extended to manifold-type processing or EPLL-type
processing to yield better results than scalar-wise estimation. The
EPLL and manifold equations should be similar to the usual methods
\cite{Zoran2011,Peyre2009} with the distinction that they are applied
on a tree structure.

\subsection{Experiments}

We apply AWGN on the detail phase coefficients of images in the DTCWT domain (the approximation
level is not affected). Applying noise on the phase distorts local
structures severely, and edges and other structures appear smeared
(Fig. \ref{fig:Phase-denoising-example}). We then apply the denoising
scheme (\ref{eq:phase_denoising}) on all detail coefficients' phases. 

In many cases, not all 9 neighbors are available, for example in the
coefficients of the finest level. In these cases, we use the model
coefficients that are present in the coefficient's neighborhood. We
emphasize that the experiments were not performed on images used for
learning. Further, the model was trained on texture datasets (Brodatz
and McGill), but was still used to denoise natural images as well.

Inspecting the results (Fig. \ref{fig:Phase-denoising-example}),
we observe that fine detailed structures, degraded severely by the
phase noise, were made more coherent, rendering the image to become more
visually appealing. We observe also the significant improvement in
the SSIM (superimposed on each figure), due to the improved structures
in the reconstructed images. These results, of images of various modalities
- both natural and textured images - demonstrate the efficacy of the
phase model in its ability to reconstruct local structures.

\renewcommand{\curww}{0.320}
\newcommand{\imphdn}[2]{\includegraphics[width=\curww\columnwidth,clip=true,trim=0 40 40 0]{phasedenoise_res/im_#1_#2}}

\renewcommand{\subs}[1]{
\subfloat[Ground truth \label{figw:phdn#1gt}]{\imphdn{#1}{gt}}~
\subfloat[Degraded \label{figw:phdn#1deg}]{\imphdn{#1}{deg}}~
\subfloat[Reconstructed \label{figw:phdn#1rec}]{\imphdn{#1}{rec}} \\
\subfloat[Degraded diff. \label{figw:phdn#1dn}]{\imphdn{#1}{diff_deg}} ~
\subfloat[Reconstructed diff. \label{figw:phdn#1dr}]{\imphdn{#1}{diff_rec}}
}
\newcommand{\subsb}[1]{
\subfloat[Ground truth \label{figw:phdn#1gt}]{\imphdn{#1}{gt}} ~
\subfloat[Degraded \label{figw:phdn#1deg}]{\imphdn{#1}{deg}} ~
\subfloat[Reconstructed \label{figw:phdn#1rec}]{\imphdn{#1}{rec}}
}
\begin{figure}[t]
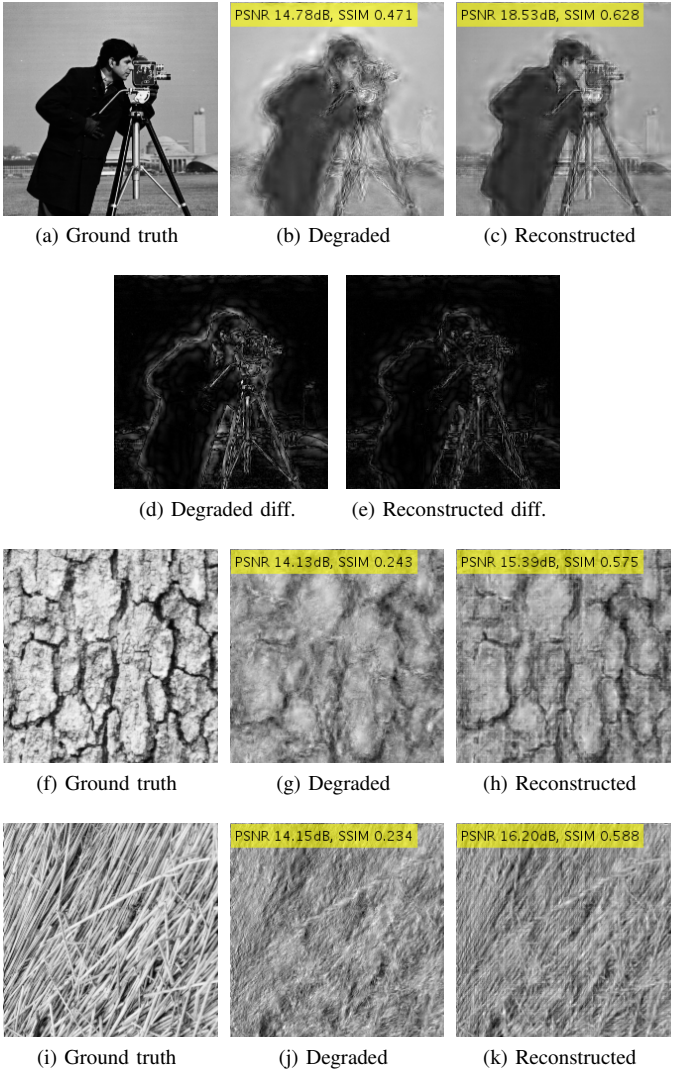
 \centering
\newcommand{\subreg}[2]{\protect\subref{figw:phdn#1#2}}

\subs{1} \\
\subsb{2} \\
\subsb{3}

\caption{Phase denoising example. The image is degraded by noisy local phase ($\sigma=2$) in its detail coefficients \subreg{1}{deg} depicts distorted local structure. The reconstructed image \subreg{1}{rec} has recovered structure, noticeably in the face and camera area. The absolute difference between the ground truth \subreg{1}{gt} and the degraded and reconstructed images, shown in \subreg{1}{dn} and \subreg{1}{dr}, respectively, illustrate that the reconstructed image has recovered some missing structure. Two additional examples are provided in the third and fourth rows: \subreg{2}{gt}, \subreg{2}{deg}, \subreg{2}{rec}: Ground truth, degraded and reconstructed images, respectively;  \subreg{3}{gt}, \subreg{3}{deg}, \subreg{3}{rec}: ground truth, degraded and reconstructed images, respectively. PSNR and SSIM values are superimposed on each image. 
\label{fig:Phase-denoising-example}}
\end{figure}

\section{Phase retrieval using the local phase model }

As a second application based on the local phase, we present a phase retrieval algorithm using
the local phase model. It is a novel, unexpected, idea to show that local phase models can
be used in global phase restoration, as is considered in the phase
retrieval problem. This important property is demonstrated in Fig.
\ref{fig:local-phase-aids} and further elaborated in Appendix \ref{app:localphase}.
In the Figure, we observe that using the ground truth local phase obtained from the wavelet
graph, restores fine-details of edge structure in images with global phase
distortion. This apparent relationship between the local and global
phase is the subject of this section, wherein we exploit the local
phase model for the benefit of the phase retrieval problem, in which case
the global phase is unknown.

\renewcommand{\figim}[2]{\includegraphics[width=0.30\columnwidth,clip=false]{phaseret_proj/i#1_#2}}

\renewcommand{\subs}[1]{
\subfloat[Ground truth\label{figw:gt#1}]{\figim{#1}{gt}} ~
\subfloat[Noisy global phase\label{figw:deg#1}]{\figim{#1}{deg}}~
\subfloat[Local phase projection\label{figw:proj#1}]{\figim{#1}{proj}}
}
\begin{figure}
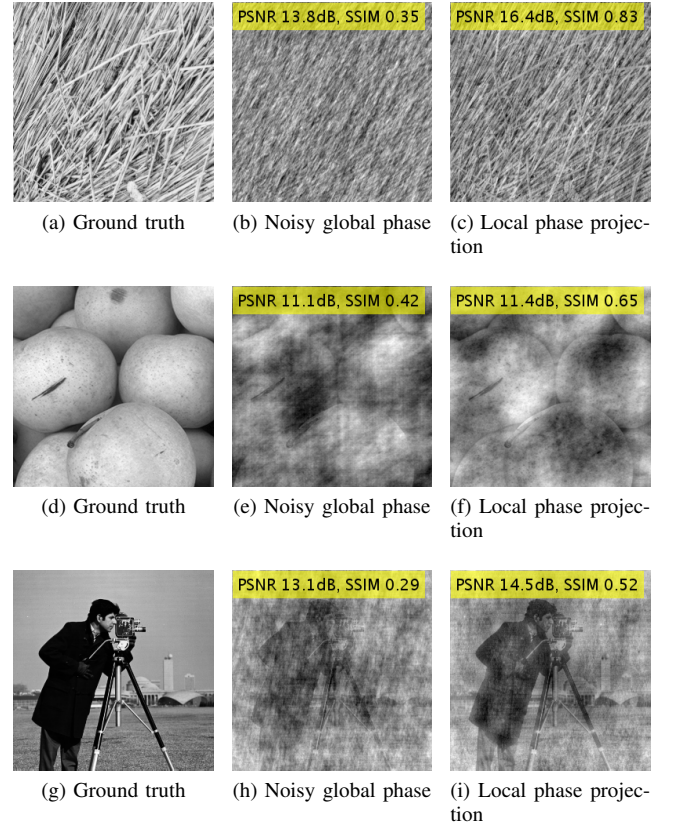
 \centering
\newcommand{\expa}{1}
\newcommand{\expb}{2}
\newcommand{\expc}{3}
\newcommand{\subreb}[2]{\protect\subref{figw:#1#2}}

\subs{\expa} \\
\subs{\expb} \\
\subs{\expc}

\newcommand{\callii}[1]{\subreb{#1}{1}, \subreb{#1}{2} and \subreb{#1}{3}}
\caption{Local phase facilitates global phase reconstruction. Examples of natural texture, natural image and man-made are shown in the left column (\callii{gt}, respectively). Gaussian noise with $\sigma=1.5$ is applied to the global phase of the ground truth images (\callii{deg}, respectively). Images with restored (projected) ground truth local phase are shown in \callii{proj}, respectively. PSNR and SSIM values are superimposed on each image. The images in the right column illustrate that the projection of the local phase improves the edge structure of the image with distorted global phase.
\label{fig:local-phase-aids}}
\end{figure}

Phase retrieval is the process of reconstructing an image in the space
domain, $I\left(x,y\right)$, given only its Fourier magnitude, $|\hat{I}(u,v)|$.
That is, the phase, $\angle\hat{I}\left(u,v\right)$ needs to be recovered
and the image thereby reconstructed. This is commonly solved iteratively
by imposing known constraints in the space domain, e.g. non-negativity
and support, and then imposing complementary constraints in the frequency
domain, usually the known magnitude. This process is iterated until
the image is restored, or a given error criterion is achieved.

The error measure commonly used for phase retrieval is the error in
the frequency domain \cite{Osherovich2011}:
\begin{equation}
\epsilon_{t}=\Vert |\hat{I}|-|\hat{I}_{t}|\Vert _{2}^{2},\label{eq:FourierERror}
\end{equation}
where $\hat{I}_{t}$ is the Fourier transform of the restored image
in the $t$th iteration. A well-known algorithm used for phase retrieval is Fienup's hybrid
input-output (HIO) algorithm \cite{Fienup1982,Shechtman2014a}, that will be used as a benchmark in our experiments.

Phase retrieval is still a challenging problem in imaging and image processing. There has been considerable advancement in solving the problem for 1D or sparse signals \cite{Shechtman2014a, Bendory2017, Jaganathan}, or by using methods that require matrix lifting, which renders the algorithms to become inefficient even in the case of moderately sized 2D images \cite{Chandra}. 

When we turn to solving the phase retrieval problem for 2D, non-sparse, moderately-sized images, we observe that HIO remains a preferred choice \cite{Chandra}. It is based on alternating projections, which does not entail working in higher dimensions. Using the HIO algorithm as our benchmark serves two goals: first, it is a widely-used phase retrieval algorithm. Second, it allows us to introduce a simple method that improves it, which does not necessitate any new assumptions (such as sparsity or maximal signal dimension).

To introduce the local phase model in phase retrieval, we modify HIO by adding local phase denoising steps. We first model the local phase as noisy with with variance $\sigma_{n}^{2}\left(i\right)$
where $i$ is the iteration number, and apply the phase denoising
scheme (\ref{eq:phase_denoising}) every $N_{est}$ iterations of
the HIO algorithm. The complete local phase-based phase retrieval
algorithm (denoted local phase HIO or LPHIO) is denoted by Alg. \ref{alg:The-local-phase}.

\begin{algorithm}
LPHIO phase retrieval algorithm. Inputs: Fourier magnitude $\left|\hat{I}\right|$,
space-domain support $S$, phase model $\Phi$, number of iterations
$T$, local phase estimation frequency $N_{est}$, expected noise
per iteration, $\sigma_{n}^{2}\left(i\right),i=0,...,T$ and HIO scaling
parameter $\beta$.

Denote by $\hat{\phi}^{i}$ the phase estimation in the $i$th iteration.
Assume $\hat{\phi}^{i}$ is initialized as uniformly distributed random
phase ($\mathcal{U}\left[-\pi,\pi\right]$).

For $i=1,...,T$:
\begin{enumerate}
\item Estimate image using the known Fourier magnitude and phase estimate:
$I^{i}\left(x,y\right)\leftarrow\text{Real}\left\{ \mathcal{F}^{-1}\left\{ \left|\hat{I}\right|e^{j\hat{\phi}^{i-1}}\right\} \left(x,y\right)\right\} .$
\item $V\left(x,y\right)\leftarrow\left\{ x,y:\,\bar{S}\left(x,y\right)\vee\left\{ I^{i}\left(x,y\right)<0\right\} \right\} $
\item $\hat{I}^{i}\left(x,y\right)\leftarrow\begin{cases}
I^{i}\left(x,y\right), & V\left(x,y\right)=0\\
I^{i-1}\left(x,y\right)-\beta\cdot I^{i}\left(x,y\right), & V\left(x,y\right)=1
\end{cases}$
\item If $mod\left(i,N_{est}\right)=0$, perform local phase estimation:\label{enu:localphase_step}
\begin{enumerate}
\item Extract the top-left $N\times N$ image $I_{tl}^{i}\left(x,y\right)$
from $\hat{I}^{i}\left(x,y\right)$. 
\item Estimate local phase for $I_{tl}^{i}\left(x,y\right)$ given $\Phi$
and $\sigma_{n}^{2}\left(i\right)$ to yield $\hat{I}_{tl}^{i}\left(x,y\right)$
using (\ref{eq:phase_denoising}).
\item Set the top-left $N\times N$ image of $\hat{I}^{i}\left(x,y\right)$
to $\hat{I}_{tl}^{i}\left(x,y\right)$.
\end{enumerate}
\item $I^{i+1}\leftarrow\hat{I}^{i}$
\item $\hat{\phi}^{i+1}\leftarrow\angle\mathcal{F}\left\{ I^{i}\left(x,y\right)\right\} $
\end{enumerate}
\caption{The local phase HIO (LPHIO) algorithm\label{alg:The-local-phase}.
The local phase estimation step \ref{enu:localphase_step} is evaluated
every $N_{est}$ iterations. The rest of the algorithm is identical
to HIO.}
\end{algorithm}

To train the phase model, we use, as described earlier, $384$ $N\times N$-sized
imaged from Brodatz and McGill datasets (cropped) with $K=10$ Gaussian
mixture components. Estimation of the phase noise variance in each
iteration is a subject of further discussion, as the wrapping of the
phase discourages using usual noise estimation techniques; Euclidean
distances (as measured by the variance) might not reflect the true
noise in the coefficients. Instead, we use the following noise model:
$\sigma_{n}^{2}\left(i\right)=a\cdot\exp\left\{ -b\cdot i/T\right\}$,
where $a$ is the initial noise estimate, $b$ determines the rate
of decay and $T$ is the number of iterations. We use the same values
of $a$, $b$ and $T$ in all experiments. 

\begin{figure}
\noindent \centering{}\includegraphics[width=1\columnwidth]{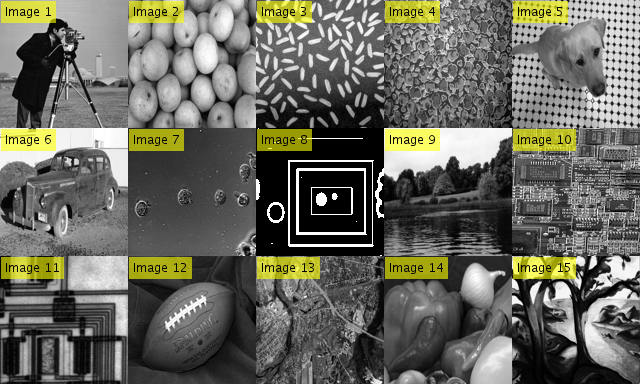}\caption{Image dataset used in evaluation of phase retrieval: the images used were
natural, man made and structured texture.\label{fig:Image-dataset-for-phaseret}}
\end{figure}
The evaluation of the algorithm is performed as follows: images
of size $N\times N=128\times128$ that are padded to size $2N$ along
each dimension are used. The $D=15$ images are of different modalities
(texture, structured texture man-made and natural images; Fig \ref{fig:Image-dataset-for-phaseret}). We note that these are arbitrary images taken from {\sc Matlab}'s image database.
The algorithm is evaluated using the following parameters: $T=1500$,
$N_{est}=50$, $a=2$, $b=5$. We measure the Fourier domain error
(\ref{eq:FourierERror}), the PSNR and SSIM of the restored images.
Since the restored images may be rotated by $180^{\circ}$ due to
the phase ambiguity, we measure the PSNR and SSIM w.r.t both the rotated
and non-rotated images and consider the maximal value. For each image,
we initialize the phase randomly $P_{init}=5$ times, for assuring
that the result is not due to different initializations of
the phase. The total number of experiments is $N_{exp}=D\times P_{init}=75$.
Fig. \ref{fig:phaseret_examples} shows an example of several image
results. 

\renewcommand{\figim}[3]{\includegraphics[width=0.23\columnwidth,clip=true,trim=0 25 0 0]{phaseret_res/i#1_s#2_#3}}
\newcommand{\figgraph}[2]{\includegraphics[width=0.25\columnwidth,clip=false]{phaseret_res/i#1_s#2_ferr}}

\renewcommand{\subs}[2]{
\subfloat[Ground truth\label{figw:gt#1_#2}]{\figim{#1}{#2}{gt}}~
\subfloat[LPHIO\label{figw:our#1_#2}]{\figim{#1}{#2}{our}}~
\subfloat[HIO\label{figw:hio#1_#2}]{\figim{#1}{#2}{hio}}~
\subfloat[Fourier error\label{figw:ferr#1_#2}]{\figgraph{#1}{#2}}
}
\begin{figure}
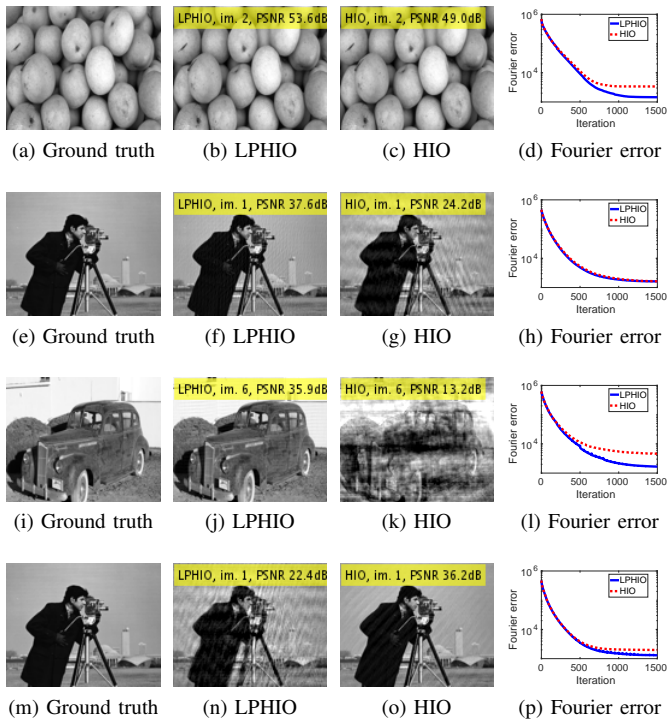
 \centering
\newcommand{\cim}{2} 
\newcommand{\cse}{2} 
\newcommand{\subre}[1]{\protect\subref{figw:#1\cim_\cse}}
\newcommand{\subreb}[3]{\protect\subref{figw:#1#2_#3}}

\subs{\cim}{\cse} \\

\renewcommand{\cim}{1} 
\renewcommand{\cse}{5} 
\subs{\cim}{\cse} \\

\renewcommand{\cim}{6} 
\renewcommand{\cse}{5} 
\subs{\cim}{\cse} \\

\renewcommand{\cim}{1} 
\renewcommand{\cse}{4} 
\subs{\cim}{\cse} 
\newcommand{\callim}[1]{\subreb{#1}{2}{2}, \subreb{#1}{1}{5}, \subreb{#1}{6}{5} and \subreb{#1}{1}{4}}
\caption{Characteristic examples of (cropped) images and phase initializations for phase retrieval using LPHIO. First column (\callim{gt}) depicts ground truth images, second column (\callim{our}, respectively) provides the results obtained by means of the proposed algorithm (LPHIO), on the third column (\callim{hio}, respectively) are the HIO results and the fourth column (\callim{ferr}, respectively) shows the Fourier domain errors, where LPHIO and HIO are shown in solid blue and dashed red lines, respectively. PSNR figures are superimposed on the reconstructed images. We observe that the Fourier error decreases with iterations. Rows 1--3 show examples of better LPHIO performance and the fourth row shows better HIO performance in terms of PSNR \subreb{hio}{1}{4} whereas the Fourier error \subreb{ferr}{1}{4} was better for LPHIO. 
\label{fig:phaseret_examples}}
\end{figure}

The $N_{exp}$ experiments are analyzed together by evaluating both
the Fourier domain error and the PSNR with respect to the ground truth. We note
that Parseval's theorem does not hold in this case, since we calculate
the Fourier domain error for coefficients of the zero-padded image
(to size $2N\times2N$), whereas we calculate the PSNR only for the
on-support image, of size $N\times N$. 

\begin{figure}
\noindent \begin{centering}
\subfloat[$d_{F}$, Fourier log-error difference\label{fig:phrethista}]{\includegraphics[bb=50bp 30bp 360bp 300bp,clip,width=0.5\columnwidth]{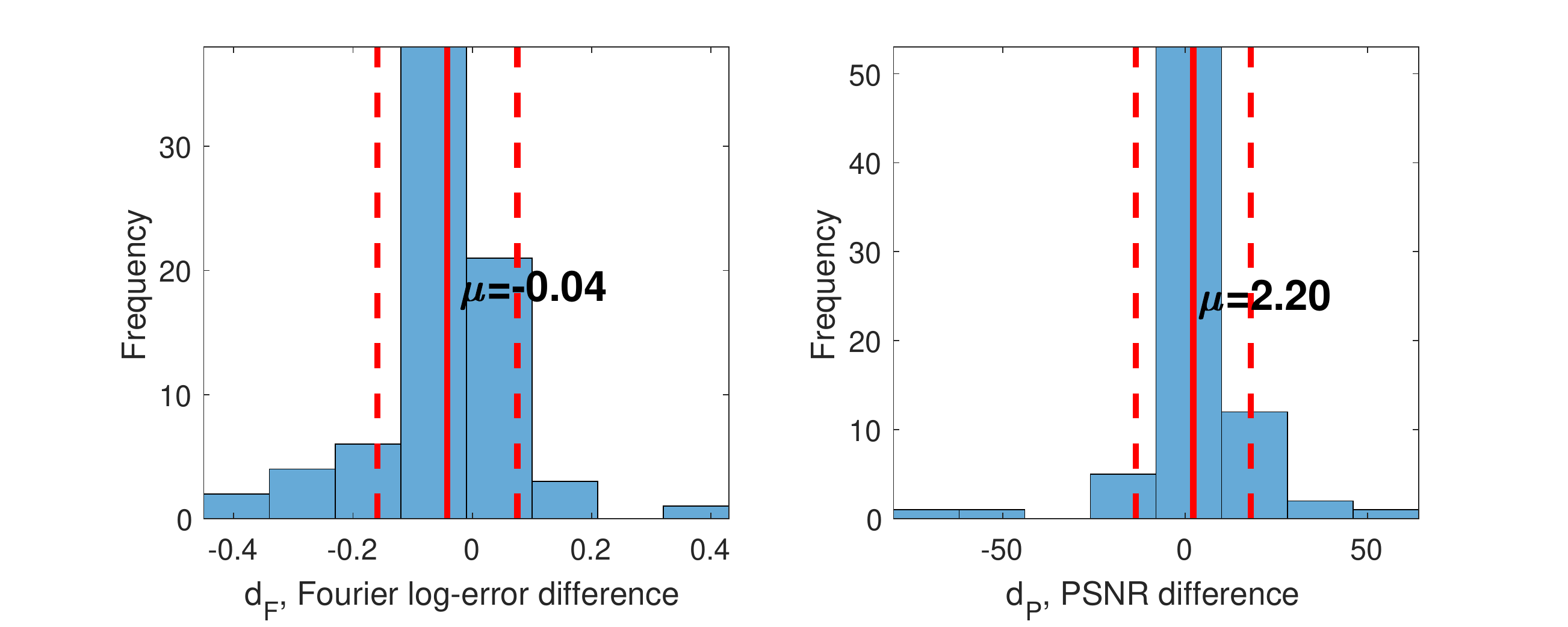}}\subfloat[$d_{P}$, PSNR difference\label{fig:phrethista-1}]{\includegraphics[bb=380bp 30bp 690bp 300bp,clip,width=0.5\columnwidth]{phaseret_hist_all_review1}}
\par\end{centering}
\caption{Comparison of LPHIO and HIO phase retrieval. Negative values indicate
lower figures for LPHIO compared with HIO. The mean and standard deviation
are highlighted by solid and dashed lines, respectively. The mean
for the Fourier domain error distance \protect\subref{fig:phrethista}
was negative and the mean for the PSNR difference \protect\subref{fig:phrethista-1}
was positive, indicating that on average, the LPHIO yielded better
results in both Fourier and object domains. \label{fig:Evaluating-LPHIO-vs.}}
\end{figure}

Two measurements,
the Fourier domain log-error difference ($d_{F}$) and the PSNR difference
($d_{P}$), are shown for each image, when using the proposed algorithm (LPHIO)
vs. standard HIO (Fig. \ref{fig:Evaluating-LPHIO-vs.}). These measurements
are defined as:
\begin{align*}
d_{F}\left(k\right) & \triangleq\log\epsilon_{t}^{LPHIO}\left(k\right)-\log\epsilon_{t}^{HIO}\left(k\right)\\
d_{P}\left(k\right) & \triangleq\text{PSNR}\left(I_{k},\hat{I}_{k}^{LPHIO}\right)-\text{PSNR}\left(I_{k},\hat{I}_{k}^{HIO}\right),
\end{align*}
where $k$ is the image index, $\epsilon_{t}^{A}$ is the Fourier
domain error for algorithm $A\in\left\{ \text{LPHIO,HIO}\right\} $,
$I_{k}$ is the ground truth $k$th image and $\hat{I}_{k}^{A}$ is
the reconstructed $k$th image using algorithm $A$. Histograms for
$d_{F}\left(k\right)$ and $d_{P}\left(k\right)$ show that on average,
the LPHIO yields lower Fourier domain error and higher PSNR compared
with HIO. LPHIO therefore yields better objective results in both metrics.

\section{Image processing scheme incorporating the phase model}

As a final example, we provide a framework for image processing using
the phase model. The phase is expressed by means of the complex wavelet
coefficients as follows:
\[
\angle x=\arctan2\left(\Im x,\Re x\right),
\]
where $\arctan2$ is the function with suitable quadrant selection.
It is observed that even in the case where a model of the phase is obtained,
a direct application of the phase model to the space domain is not
trivial. 

Consider the following problem:
\[
y=Hx+n,
\]
where $x$, $n$ and $y$ are the ground truth, independent noise
source and the degraded image, respectively. $H$ is assumed to be
a known linear operator, e.g. blur. Restoration is performed by solving
the following MAP problem:
\begin{align*}
\hat{x} & =\arg\max_{x}\P{y|x}\P x.
\end{align*}
Let $w=Wx$ denote the wavelet coefficients of $x=W^{-1}w$ and $W$
denote the wavelet transform. Then, 
\begin{align*}
\hat{x} & =\arg\min_{x}\left(-\log\P{y|w}-\log\P w\right)\\
 & =\arg\min_{x}\frac{1}{2\sigma_{n}^{2}}\left\Vert HW^{-1}w-y\right\Vert ^{2}+f\left(w\right)\\
 & \triangleq\arg\min_{x}\beta\left\Vert HW^{-1}w-y\right\Vert ^{2}+f\left(w\right).
\end{align*}
Direct application of the phase model here is challenging, since $w$
is complex and we model its phase. Instead, we have access to $\P{\angle w}$.
Denoting $a_{w}\triangleq\angle w$, we present the following formulation,
known as half quadratic splitting \cite{Zoran2011}:
\begin{equation}
L=\beta\left\Vert HW^{-1}w-y\right\Vert ^{2}+\alpha\left\Vert a_{w}-z\right\Vert ^{2}-\log\P z.\label{eq:halfquadsplit_formulation}
\end{equation}
Problem (\ref{eq:halfquadsplit_formulation}) yields the best match
of $w$ to both minimization of the likelihood term and maximization
of the probability of the phase when $\alpha\rightarrow\infty$. The
intermediate term, $\left\Vert a_{w}-z\right\Vert ^{2}$, acts as
a surrogate that links the phase prior with the signal fidelity. The
standard method of solving this problem is by iterating with increasing
values of $\alpha$, where in each iteration, the minimization problem of $L$ for $w$ given $z$ is solved first, and the second stage
is the alternate minimization ($z$ given $w$). 

The first problem entails solving
\[
\hat{a}_{w}=\arg\min L_{1}:=\beta\left\Vert HW^{-1}w-y\right\Vert ^{2}+\alpha\left\Vert a_{w}-z\right\Vert ^{2}
\]
for $a_{w}$, where $a_{w}=\angle w$. The intuitive meaning of this problem
is to find the best coefficients for the fidelity term that have the
local phase given by $z$. This problem is approximated by iterative
projections:
\begin{align*}
\hat{w} & :=\hat{w}-\lambda\nabla_{\hat{w}}\left[\left\Vert HW^{-1}\hat{w}-y\right\Vert ^{2}\right]\\
\angle\hat{w} & :=z,
\end{align*}
until convergence. 

The second problem requires solving
\[
\hat{z}=\arg\min L_{2}:=\alpha\left\Vert a_{w}-z\right\Vert ^{2}-\log\P z.
\]
This kind of problem is now detached from the meaning of the coefficients
(phase) and was solved approximately elsewhere (e.g. \cite{Zoran2011}).
The approximation entails calculating the posterior of the Gaussian
components, $\hat{k}=\arg\max_{k}\P{k|z}$, and choosing the distribution
as a single Gaussian of the $\hat{k}$'th Gaussian in the mixture.
Then, a MAP estimate can be formulated in close form:
\[
\hat{z}=\left(\Sigma_{\hat{k}}+\alpha I\right)^{-1}\left(\Sigma_{\hat{k}}a_{w}+\alpha I\mu_{\hat{k}}\right).
\]

\section{Discussion and further work}

While the importance of phase has been investigated thoroughly in
recent decades, much of the results were of projection algorithms
to some constraints set or likelihood models. In this work we provided
a prior model that can be used in Bayesian settings and illustrated
several applications. 

Using the prior model for phase retrieval has been demonstrated for
arbitrary images that do not belong to the domain of the trained model (i.e. textures), thereby
indicating the generality of the local phase model. It is our underlying
assumption that images are characterized by the same overall behaviour
as the learned images. This assumption renders our algorithm to be less restrictive than other
phase retrieval algorithms that assume sparsity in the object domain
\cite{Shechtman2014a}, or some strict knowledge of the true global
phase \cite{Osherovich2011}. 

In a discrete wavelet representation of images, we decompose the image several times, so that the final approximation coefficients provide little information concerning fine image details. They encode background information of lower frequencies. In this study, we did not model these coefficients, since their phase is less pronounced. However, it is interesting to provide a joint model of details and approximation coefficients, and to compare it with the proposed model.

Decomposing an image by means of wavelets is a well-known method of image analysis. However, the application of this technique as presented in this work, by locally modelling space and scale coefficients, was not considered in this form earlier. In other studies, image phase is considered to be a global property, or a property that is related only to scale. The proposed graph representation encapsulates both spatial and scale adjacencies in a unified model. 

While we presented the image in a graph structure, the model did not exploit any advanced graph properties other than the neighborhood structure. Applying graph processing techniques to our proposed representation should present an interesting topic to pursue in further work. 

Phase retrieval is in particular an interesting and challenging problem, in which the HIO algorithm is still preferred for the case of non-sparse images. The phase retrieval algorithm presents several further challenges in terms of accurate noise estimation and efficiency. The noise model
used so far was heuristic. Better results may be obtained by using
more realistic modelling and estimation of the local phase noise. The
efficiency of the phase estimation has not been optimized so far.
That can be obtained by using advanced tree-based methods, suitable
for the local phase structure as sub-trees of a wavelet decomposition
tree.

\section*{Acknowledgement}

This research was supported in part by the Israeli Chief Scientist
under the National Consortium OMEK, and by the Ollendorff Minerva Center.

\appendices

\section{Effects of global phase deviation on local frequency coefficients}\label{app:localphase}

To capture the essence of the relationship between global and local
phase, we analyze 1D signals in a discrete domain, that
can be further elaborated to deal with 2D signals. The Fourier transforms are
discrete Fourier transforms, i.e. DFT, and the convolutions are also
discrete. We use local analysis by means of the STFT for simplicity.

Let $x\left(t\right)$ denote a 1D signal of a discrete index $t$,
and $\hat{x}\left(f\right)$ its Fourier transform. Let $x'\left(t\right)$
denote a signal identical to $x\left(t\right)$ other than a small
perturbation $\eta$ in the phase of one of the components: 
\begin{align*}
\hat{x}'\left(f_{0}\right) & =\hat{x}\left(f_{0}\right)e^{j2\pi\eta}\\
x'\left(t\right) & =\mathcal{F}^{-1}\left\{ \hat{x}'\left(f\right)\right\} \left(t\right).
\end{align*}

$x'\left(t\right)$ is then given by 
\newcommand{\ton}{\frac{t}{N}}
\begin{align*}
x'\left(t\right) & =\sum_{f\neq f_{0}}\hat{x}_{f}'e^{j2\pi f\ton}+\sum_{f=f_{0}}\hat{x}_{f}'e^{j2\pi f\ton}\\
& =\tilde{x}\left(t\right)+\hat{x}_{f_{0}}e^{j2\pi f_{0}\ton}e^{j\eta}\\
& =x\left(t\right)+\hat{x}_{f_{0}}e^{j2\pi f_{0}\ton}\left(e^{j\eta}-1\right)\\
& \approx x\left(t\right)+\hat{x}_{f_{0}}e^{j2\pi f_{0}\ton}\cdot j\eta,
\end{align*}
where $\tilde{x}\left(t\right)$ is the signal $x\left(t\right)$
without the component in the frequency $f_{0}$. 

Next, we analyze the local frequency coefficients of $x'\left(t\right)$
via STFT: 
\begin{align*}
\hat{x}'\left(t_{0},f\right) & =\sum_{t}x'\left(t\right)w_{t-t_{0}}e^{-j2\pi f\ton}\\
& \approx\sum_{t}x\left(t\right)w_{t-t_{0}}e^{-j2\pi f\ton}+\\
& \quad+\sum_{t}\hat{x}_{f_{0}}e^{j2\pi f_{0}\ton}j\eta\cdot w_{t-t_{0}}e^{-j2\pi f\ton}\\
& =\hat{x}\left(t_{0},f\right)+\eta\cdot j\hat{x}_{f_{0}}\sum_{t}w_{t-t_{0}}\cdot e^{j2\pi\left(f_{0}-f\right)t/N}\\
& \triangleq\hat{x}\left(t_{0},f\right)+\eta\cdot j\hat{x}_{f_{0}}\cdot z\left(f;t_{0},f_{0}\right),
\end{align*}
where 
\begin{align*}
z\left(f;t_{0},f_{0}\right) & =\sum_{t}w_{t-t_{0}}\cdot e^{j2\pi\left(f_{0}-f\right)t/N}\\
& =e^{j2\pi\left(f_{0}-f\right)t_{0}/N}\cdot\sum_{t}w_{t-t_{0}}\cdot e^{j2\pi\left(f_{0}-f\right)\frac{t-t_{0}}{N}}\\
& \triangleq e^{j\beta}\cdot\sum_{t}w_{t-t_{0}}\cdot e^{j2\pi\left(f_{0}-f\right)\left(t-t_{0}\right)/N}\\
\left|z\left(f;t_{0},f_{0}\right)\right| & =\left|\sum_{t}w_{t-t_{0}}\cdot e^{j2\pi\left(f_{0}-f\right)\left(t-t_{0}\right)/N}\right|.
\end{align*}
Let us consider the above absolute value, $\left|z\left(f;t_{0},f_{0}\right)\right|$;
it does not depend on $t_{0}$ due to the summation over $t$. It
is a Fourier transform of a window function (sinc) shifted by frequency
$f_{0}$. It, therefore, decreases in absolute value in frequencies
near $f_{0}$.

Let us return to 
\begin{align*}
\hat{x}'\left(t_{0},f\right) & \approx\hat{x}\left(t_{0},f\right)+\eta\cdot\hat{x}_{f_{0}}\cdot e^{j\alpha}\left|z\left(f;t_{0},f_{0}\right)\right|,
\end{align*}
where $e^{j\alpha}=je^{j\beta}=e^{j\left(\pi/2+\beta\right)}$ is
some phase addition. We observe that the global phase shift of a frequency
$f_{0}$ by a quantity $\eta$ is expressed approximately in the local
frequency domain via an addition that is scaled by the phase shift,
$\eta$ and the global frequency component $\hat{x}_{f_{0}}$. Further,
it is localized in the local frequency $f_{0}$ which is the same
as the global frequency, and its localization depends on the width
of the windowing function $w_{t-t_{0}}$. It does not depend on the
spatial location, $t_{0}$. 

This analysis shows that a shift (e.g. distortion) in global phase,
expressed by the parameter $\eta$ for some frequency $f_{0}$, degrades
the local coefficients with the same frequency $f_{0}$ by a linear
addition (scaled by $\eta$) that affects both magnitude and phase
of the local coefficient. 

This relationship does not indicate that simply enforcing the correct
local phase will necessarily have the effect of minimizing the distortion
parameter $\eta$. We, therefore, resort to the following demonstration;
let us now consider an edge signal $x\left(t\right)$:
\[
x\left(t\right)=\begin{cases}
1, & t\geq0\\
-1, & t<0.
\end{cases}
\]
It has zero global phase. Further, assume that we obtain $x'(t)$
by applying a phase deviation $\eta$ and inspect the local frequencies
of $\hat{x}'\left(t_{0},f\right)$ at $t_{0}=0$. Locally we observe
the same edge structure with zero phase. In this case, we have:

\begin{align*}
\hat{x}'\left(t_{0},f\right) & \approx\left|\hat{x}\left(t_{0},f\right)\right|e^{j\angle\hat{x}\left(t_{0},f\right)}+\eta\cdot\left|\hat{x}_{f_{0}}\right|\cdot\\
& \quad\cdot\left|z\left(f;t_{0},f_{0}\right)\right|e^{j\left(\alpha+\angle\hat{x}_{f_{0}}\right)}\\
& =\left|\hat{x}\left(t_{0},f\right)\right|+\eta\cdot\left|\hat{x}_{f_{0}}\right|\left|z\left(f;t_{0},f_{0}\right)\right|e^{j\alpha},
\end{align*}
where the only complex term is $e^{j\alpha}=je^{j\beta}$, where $\beta$
is a phase shift caused by the shifted spatial window function. We
observe that in this case, enforcing the correct local phase will
necessarily lead to $\eta=0,$ i.e. the correct global phase as well.

\bibliographystyle{unsrt}
\bibliography{library}

\end{document}